\RequirePackage{fix-cm}
\documentclass[twocolumn]{svjour3}          
\smartqed  
\usepackage{graphicx}
\usepackage[utf8]{inputenc}
\usepackage{times}
\usepackage{fullpage}
\usepackage{eufrak}
\usepackage{amssymb}
\usepackage{amsmath}

\usepackage{amsthm}
\usepackage{tikz}
\usepackage{float}
\usepackage{pgfplots}
\usetikzlibrary{arrows.meta}
\usepackage{xcolor}
\usepackage[numbers,sort]{natbib}
\usepackage[backref,pageanchor=true,plainpages=false, pdfpagelabels, bookmarks,bookmarksnumbered,
]{hyperref}
\usepackage{subfigure}

\usepackage[letterpaper,twoside,vscale=.8,hscale=.75,nomarginpar,hmarginratio=1:1]{geometry}

\newtheorem*{remark*}{Remark}

\newcommand{\indep}{\raisebox{0.05em}{\rotatebox[origin=c]{90}{$\models$}}}
\newcommand{\model}{\textit{pure 1-factor measurement model\color{white}.\color{black}}}
\newcommand{\modelp}{\textit{pure 1-factor measurement models\color{white}.\color{black}}}
\newcommand{\triple}{\textit{pure triple}}

\DeclareMathOperator{\E}{\mathbb{E}}
\pgfplotsset{compat=1.14}

\makeatletter
\newcommand{\printfnsymbol}[1]{%
  \textsuperscript{\@fnsymbol{#1}}%
}
\makeatother
\begin{document}

\title{{Causal Clustering for 1-Factor Measurement Models on Data with Various Types}
}


\author{Shuyan Wang         
}


\institute{Shuyan Wang \at
              Carnegie Mellon University \\
              \email{shuyanw@andrew.cmu.edu}           
}

\date{Received: date / Accepted: date}

\maketitle

\begin{abstract}
The tetrad constraint is a condition of which the satisfaction signals a rank reduction of a covariance submatrix and is used to design causal discovery algorithms that detects the existence of latent (unmeasured) variables, such as FOFC \cite{Kummerfeld:2016:CCM:2939672.2939838}. Initially such algorithms only work for cases where the measured and latent variables are all Gaussian and have linear relations (Gaussian-Gaussian Case). It has been shown that a unidimentional latent variable model implies tetrad constraints when the measured and latent variables are all binary (Binary-Binary case). This paper proves that the tetrad constraint can also be entailed when the measured variables are of mixed data types and when the measured variables are discrete and the latent common causes are continuous, which implies that any clustering algorithm relying on this constraint can work on those cases.  Each case is shown with an example and a proof.  The performance of FOFC on mixed data is shown by simulation studies and is compared with some algorithms with similar functions.
\keywords{Causal Discovery \and Latent Variable}
\end{abstract}

\section{Introduction}
\label{intro}
Tetrad constraints is a condition of which the satisfaction signals a rank reduction of a covariance submatrix and is used to design causal discovery algorithms that detects the existence of latent (unmeasured) variables, such as FOFC  \cite{Kummerfeld:2016:CCM:2939672.2939838}.   Initially such algorithms only work for cases where the measured and latent variables are all Gaussian and have linear relations (Gaussian-Gaussian Case).  It has been shown that the tetrad constraints also hold for the case where the measured variables are all binary and they are only connected with one binary latent common cause (Binary-Binary case)\cite{DBLP:journals/corr/abs-1301-2263}. Here we prove that the tetrad constraint can also be entailed when the measured variables are of mixed data types and when the measured variables are discrete and the latent common causes are continuous, which implies that any clustering algorithm relying on this constraint can work on those cases.   In section 2, we briefly introduce the tetrad constraint and FOFC as an example of algorithms designed based on it.  In section 3, we describe each case 
for which tetrad constraint can work with a proof and an example.  In section 4, we discribe the performance of FOFC on simulation studies: in section 4.1, we describe the performance of FOFC on simuated data corresponding to cases in section3; in section 4.2 we compare FOFC using rank correlation and tetrachoric correlation; in section 4.3 we compare FOFC with MGM and MGM-FCI-MAX, two algorithms studying the structure of mixed data.

\section{FindOneFactorClusters}
\label{fofc1}
We use directed acyclic graphs to represent causal relations between variables.  Given a directed edge in the graph, a node is defined as \textbf{parent} if the edge comes from it; the \textbf{child} of the directed edge is the node that it goes into.  A \textbf{directed path} is a sequence of connected edges where the source of the latter one is the target of the first one.  Given a directed path, the parent of the first edge is the \textbf{ancestor} of the child of the last edge and the target of the last edge is the \textbf{descendant} of its \textbf{ancestor}.  Using a directed graph to represent causal relation, each edge represents one direct causal relation relative to the variables in the graph, with the direction from the cause (parent) to the effect (child).  We say a causal relation is direct relative to a set of variables $S$ if that causal relation cannot be interrupted by interfering any subset of $S$.

We assume the joint probability distribution on the variables respects the Markov condition, i.e., all variables conditioned on their parents are independent from the set of all of their non-descendants.  We also assume Faithfulness, which states that all conditional independence relations that hold in the population are consequences of the Markov condition from the underlying true causal graph.

 A \textbf{trek} between two variables, $X$ and $Y$, is defined as either a directed path between $X$ and $Y$, or two directed paths starting from a common third variable  $Z$, the two paths intersecting only at $Z$, with one path ending at $X$ and another at $ Y$. 

\begin{figure}
    \centering
    \includegraphics[width=\linewidth]{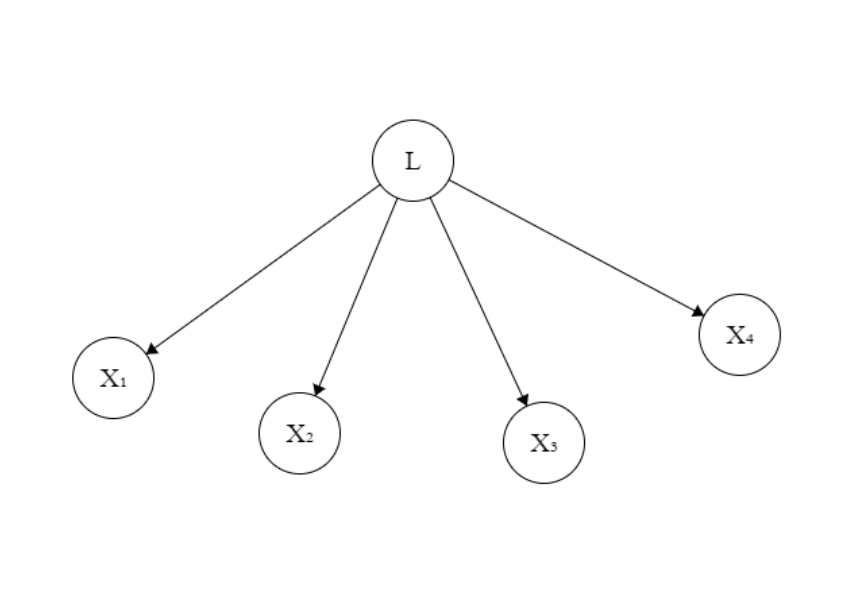}
    \caption{$L ,X_1 , X_2 , X_3$ and $ X_4$ are binary}
    \label{fig:clusterM}
\end{figure}

The FindOneFactorClusters (FOFC) algorithm aims to find the \textit{pure 1-factor measurement models}, where each measured variable shares one latent cause and there are no other causal connections between the measured variables \cite{Kummerfeld:2016:CCM:2939672.2939838}. A set of variables \textbf{\textit{V}} is causally sufficient when every cause of any two variables in \textbf{\textit{V}} is also in \textbf{\textit{V}}. Given a set of measured indicators \textbf{\textit{O}}, and a causally sufficient set of variables \textbf{\textit{V}} containing \textbf{\textit{O}} s.t. no strict subset of \textbf{\textit{V}} containing \textbf{\textit{O}} is causally sufficient, then a \textit{pure 1-factor measurement models} for \textbf{\textit{V}} is a measurement model in which each observed indicator has at most one latent parents, no observed parents, and no correlated errors.  For instance, figure \ref{fig:clusterM} is a \textit{pure 1-factor measurement model} where $X_1$ to $X_4$ are measured variables and $L$ is latent. 

If both the measured and latent variables are Gaussian and each measured variable can be written as a linear function of the latent variable plus a noise independent from other measured and latent variables,  we can derive the \textit{vanishing tetrad constraint}
\begin{center}
    $Cor(X_i, X_j)Cor(X_w, X_h)=Cor(X_w, X_j)Cor(X_i, X_h)$  
\end{center}  

among any four measured variables.  In fact, any three measured Gaussian variables in a \textit{pure 1-factor measurement model} entail a vanishing tetrad with another Gaussian variable no matter how they are partitioned.  In this case, the three measured variables are called a \textit{pure triple}.  

FOFC finds a \textit{pure 1-factor measurement model} by first identifying all \textit{pure triples}.  Then for every \textit{pure triple} the algorithm expands each triple into a cluster by merging triples with overlapping variables.  Finally it outputs the cluster with the largest size, deletes all the clusters that have overlapping variables with the largest one then outputs the largest one that remains.\cite{CW10.2307/4140591}

\section{Working Cases and Proofs}

\subsection{Binary-Binary}

Consider a case of Figure \ref{fig:clusterM}, where both the latent and measured variables are binary.  According to the lemma in\cite{10.5555/534975} \cite{DBLP:journals/corr/abs-1301-2263}:
\begin{lemma}
If $A$, $B$ and $C$ are all binary variables, $A\indep C | B$ \footnote{$A\indep C | B$ denotes that $A$ and $C$ are independent conditioning on $B$} if and only if  $Cor(A,C) = Cor(A,B)Cor(B,C)$.
\end{lemma}
for a group of binary variables sharing a single common binary cause, \textit{three vanishing tetrad constraints are satisfied:}
\begin{center}
    $Cor(X_1, X_2)Cor(X_3, X_4)=Cor(X_1, X_3)Cor(X_2, X_4)=Cor(X_1, X_4)Cor(X_2, X_3)$  
\end{center}
As long as there are at least four measured binary variables sharing the same unique binary latent common cause, and no measured variables are caused by other measured variables, FOFC will find that any three of them is a \textit{pure triple} and manage to merge the triples into a pure cluster.

\subsection{Continuous-Mixed}

For a case where the latent and some measured variables are Gaussian while others are binary, it is often assumed that the binary variable is generated by partitioning the values of the Gaussian  variable into two categories.  For the convenience of calculation, we assume all Gaussian variables are standardized.  For instance, for a binary variable $V_i$ which can take value zero or one, we assume that it is generated by a standard normal variable $X_i$ such that $V_i = 0$ if $X_i < S_i$, and $V_i = 1$ otherwise.   It will be shown in later sections that, for two binary variables, $V_i$ and $V_j$, that are generated in this way, their joint distribution is decided by the joint distribution between $X_i$ and $X_j$ (see \ref{fig:clusterCD}),  the Gaussian variables that generates each of them. Therefore, the binary variable can be viewed as a child of a Gaussian variable and four such variables form a \textit{pure 1-factor measurement model}.  If only one of the measured variables is binary and the other three are linearly related to the latent, then the measured variables still satisfies \textit{vanishing tetrad constraints}, but in general this constraint is violated.  However, in simulations with mixed types FOFC identifies clusters with two or more continuous (and linear) and two or more binary variables most of the time.  This section will introduce why tetrad constraints are not entailed when there are two or more binary measured variables, and describe conditions where tetrad constraints approximately hold when there are two or more binary variables. 

\subsubsection{One Binary Variable}
For a binary variable $V_i$ generated as described above, its covariance with a Gaussian $X_i$ :
\begin{center}
    $Cov(V_i, X_i)=\int_{S_i}^\infty x\phi(x)dx = (\sqrt{2\pi})^{-1}exp({-S_i^2/2})$
\end{center}
where $\phi$ is the pdf of standard normal variable.  
For any other $X_j$ that forms a \textit{pure 1-factor measurement model}, we have:
\begin{center}
    $Cov(V_i,X_j)= Cor(X_i, X_j)Cov(V_i,X_i)$
\end{center}
Since $X_i$ forms a \textit{pure triple} with other measured Gaussian variables, replacing $X_i$ with $V_i$ does not change the pure property of those triples, i.e., adding another variable to the triple creates a vanishing tetrad regardless of the partition.  Therefore, FOFC is able to identify them and correctly cluster $V_i$ with other measured Gaussian variables that form a \textit{pure 1-factor measurement model} with $X_i$. 

We can generalize this case: when $V_i$ is some function of $X_i$ and some independent noise $e$, i.e., $V_i=f(X_i,e_v)$ while other variables in the model are still linear Gaussian, the vanishing tetrad is still entailed by \textit{pure triple}.  If we denote the latent common cause as $L$ such that $X_j=a_j L+e_j$ for all $j$, we have:
\begin{equation} 
\label{eq1}
\begin{split}
Cov(V_i, X_j) & = \E(V_i X_j)-\E(V_i)\E(X_j)\ \\
 & = \E(f(X_i,e_v)(a_j L+e_j)) \ \\
 & = a_j \E(L f(X_i,e_v))\
\end{split}
\end{equation}
Then the tetrad constraint holds for any partition:
\begin{center}
    $Cov(V_i,X_j)Cov(X_h,X_w)=Cov(V_i,X_h)Cov(X_j,X_w)=a_j a_h a_w \E(L f(X_i,e_v))$
\end{center}

\subsubsection{More than One Binary Variable}

When several measured variables are binary while the latent variable is continuous, the measured variables would fail to form \textit{pure triples} that entail the tetrad constraints.  Assume two of the measured variables, $V_i$ and $V_j$, are binary and are generated by two standard Gaussian variables $X_i$ and $X_j$ as described before, and let $\Psi$ be the cdf of the joint standard bivariate normal distribution between $X_i$ and $X_j$ and $\rho_{ij}$ the correlation between $X_i$ and $X_j$, we get the covariance between $V_i$ and $V_j$:
\begin{equation} 
\label{eq1}
\begin{split}
&Cov(V_i, V_j)\ \\
& = \E(V_i V_j)-\E(V_i)\E(V_j)\ \\
 &=\mathbb{P}(V_i=1, V_j=1)-\mathbb{P}(V_i=1)\mathbb{P}(V_j=1)\ \\
 &=\mathbb{P}(X_i>S_i, X_j>S_j)-\mathbb{P}(X_i>S_i)\mathbb{P}(X_j>S_j)\ \\
 & = \Psi(-S_i,-S_j, \rho_{ij})-\Psi(-S_i,-S_j, 0) \
\end{split}
\end{equation}

Theoretically, FOFC is not entailed to build the right cluster, but the accuracy of FOFC clustering of measured variables with mixed types is too high to be chance.
Similar to the last case, we can generalize this case: when $V_i$ is some function of $X_i$ and some noise $e$, i.e., $V_i=f(X_i,e_{v_i})$ and  $V_j$ is some function of $X_j$ and some noise $e$, i.e., $V_j=g(X_j,e_{v_j})$ while other variables in the model are still linear Gaussian, one tetrad constraint is entailed.  If we denote the latent common cause as $L$ such that $X_h=a_h L+e_h$ for all $h$, we have:
\begin{center}
     $Cov(V_i, X_h) = a_h \E(L f(X_i,e_{v_i}))$
\end{center}
\begin{center}
    $Cov(V_j, X_h) = a_h \E(L g(X_i,e_{v_j}))$
\end{center}
Then the tetrad constraint holds for the partition where the correlation between $V_i$ and $V_j$ is not included:
\begin{center}
    $Cov(V_i,X_w)Cov(V_j,X_h)=Cov(V_i,X_h)Cov(V_j,X_w)=a_h a_w \E(L f(X_i,e_{v_i}))\E((L g(X_i,e_{v_j}))$
\end{center}

\subsection{Continuous-Binary}
\begin{figure}[H]
    \centering
    \includegraphics[width=\linewidth]{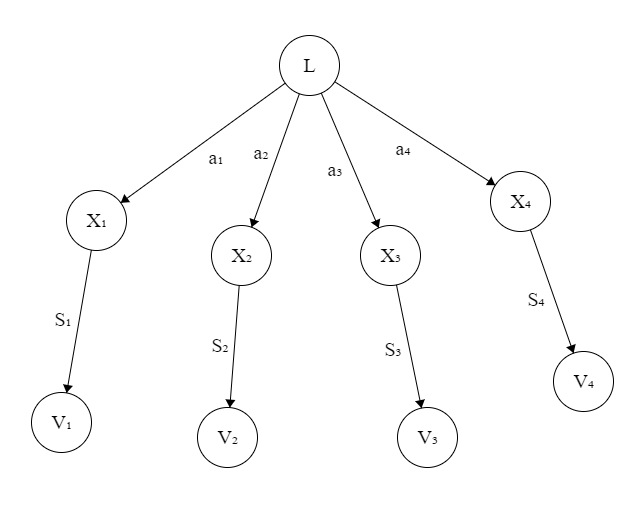}
    \caption{$L ,X_1 , X_2 , X_3$ and $ X_4$ are Gaussian, $V_1 , V_2 , V_3$ and $ V_4$ are binary}
    \label{fig:clusterCD}
\end{figure}

For the case when the latent variable is Gaussian and measured variables are binary, we assume that the measured variables are generated with the same mechanism as described in the last section. As shown in figure \ref{fig:clusterCD}, the mediate variables $X_1$ to $X_4$ forms a \model.  Each mediate Gaussian variable can be written as a linear function of the latent variable $L$:
\begin{center}
    $X_i = a_i L + e_i$
\end{center}
where $a_i$ is constant and $e_i$ represents the extra cause of $X_i$ that is independent from $L$.  As mentioned before, the measured binary variable $V_i =1$ if $X_i > S_i$ for some constant $S_i$ and $V_i = 0$ otherwise. From the section 3.2.2 we know that the covariance between binary variables generated in this way does not reliably allow any three measured variables to form a \triple, and FOFC is not entailed to work in this case.  However, simulation tests show it is nonetheless very accurate. This phenomenon is going to be discussed in the next two sections, from median dichotomy ($S_i = 0$) to non-median dichotomy.

\subsubsection{Median Dichotomy}
Suppose a binary variable $V_i$ is generated from a median dichotomy, i.e., $S_i =0$.  When every measured variables is generated from a median dichotomy, the covariance matrix of the measured data behaves quite similarly to a covariance matrix of the  Gaussian variables.  Given two standard Gaussian variables $u$ and $v$ with correlation $\rho$, one property of bivariate normal distribution formed by $u$ and $v$ is that:
\begin{center}
    $\dfrac{\partial \Psi (u,v,\rho)}{\partial \rho} = \psi (u,v,\rho)$
\end{center}
where $\psi(x,y,z)$ is the pdf of standard bivariate normal distribution\cite{BGCdoi:10.1080/03610926.2011.611316}.
Then the covariance between two measured binary variables is:
\begin{equation} 
\label{eq1}
\begin{split}
Cov(V_i, V_j)
 & = \Psi(-S_i,-S_j, \rho_{ij})-\Psi(-S_i,-S_j, 0) \ \\
 & = \int_0^{\rho_{ij}}\psi (-S_i,-S_j,r)dr
\end{split}
\end{equation}
When $S_i=S_j=0$ the covariance can be written as :
\begin{equation} 
\label{eq3}
\begin{split}
Cov(V_i, V_j)
 &=\int_0^{\rho_{ij}}\psi (0,0,r)dr\ \\
 &=\int_0^{\rho_{ij}}\dfrac{1}{2\pi\sqrt{1-r^2}}dr\ \\
 &=\dfrac{1}{2\pi}arcsin(\rho_{ij})\ \\
 & \approx \dfrac{1}{2\pi}(\rho_{ij}+\dfrac{\rho_{ij}^3}{6})
\end{split}
\end{equation}
The last step is based on the Taylor expansion of $arcsin$.
\textbf{Equation(3)} indicates that the covariance between the two binary variables with median dichotomy is proportional to the correlation between the two mediate Gaussian variables.  If $|\rho| \leq 0.65$, the difference between $arcsin(\rho)$ and $\rho$ is less than 10\% 
\cite{CW10.2307/4140591}.  Since the covariance between the measured binary variables is approximately proportional to the correlation between the mediate Gaussian variables, the property that the a partition of any four mediate variables satisfies the tetrad constraints approximately holds on the measured binary variables.  In fact, the performance of FOFC on binary data with median dichotomy is almost as good as on Gaussian data.

\subsubsection{Non-median Dichotomy}

Recall that we assume the binary variable $V_i$ is generated from a standard Gaussian variable $X_i$ with a cutoff value $S_i$, such that $V_i=1$ if $X_i>S_i$ and $V_i=0$ otherwise.  This section discusses the situation where $S_i \neq 0$.  When the $|S_i| \leq 1$, FOFC is still reliable for building the right cluster. Recall the covariance between two binary variables:
\begin{align}
\begin{split}
&Cov(V_i, V_j)\ \\
 &=\int_0^{\rho_{ij}}\psi (-S_i,-S_j,r)dr\ \\
 &=\int_0^{\rho_{ij}}\dfrac{1}{2\pi\sqrt{1-r^2}}exp\{\frac{S_i^2-2r S_i S_j+S_j^2}{-2(1-r^2)}\}dr
 \end{split}
 \end{align}
 When $r$ is small and the absolute value of $S_i$ and $S_j$ are smaller than 1 , $1-r^2$ is approximately 1 and $rS_iS_j$ is apporximately 0. The covariance can be approximated:

 \begin{align}
 \begin{split}
 Cov(V_i,V_j)
 &\approx\int_0^{\rho_{ij}}\dfrac{1}{2\pi\sqrt{1-r^2}}exp\{\frac{S_i^2+S_j^2}{-2}\}dr\ \\
 &=exp\{\frac{S_i^2+S_j^2}{-2}\}\int_0^{\rho_{ij}}\dfrac{1}{2\pi\sqrt{1-r^2}}dr\ \\
 &=exp\{\frac{S_i^2}{-2}\}exp\{\frac{S_j^2}{-2}\}\dfrac{1}{2\pi}arcsin(\rho_{ij})
\end{split}
\end{align}
The final step of the approximation entails the binary variable with non-median dichotomy to satisfy the tetrad constraints.

\begin{figure*}
    \centering
    \includegraphics[width=\textwidth]{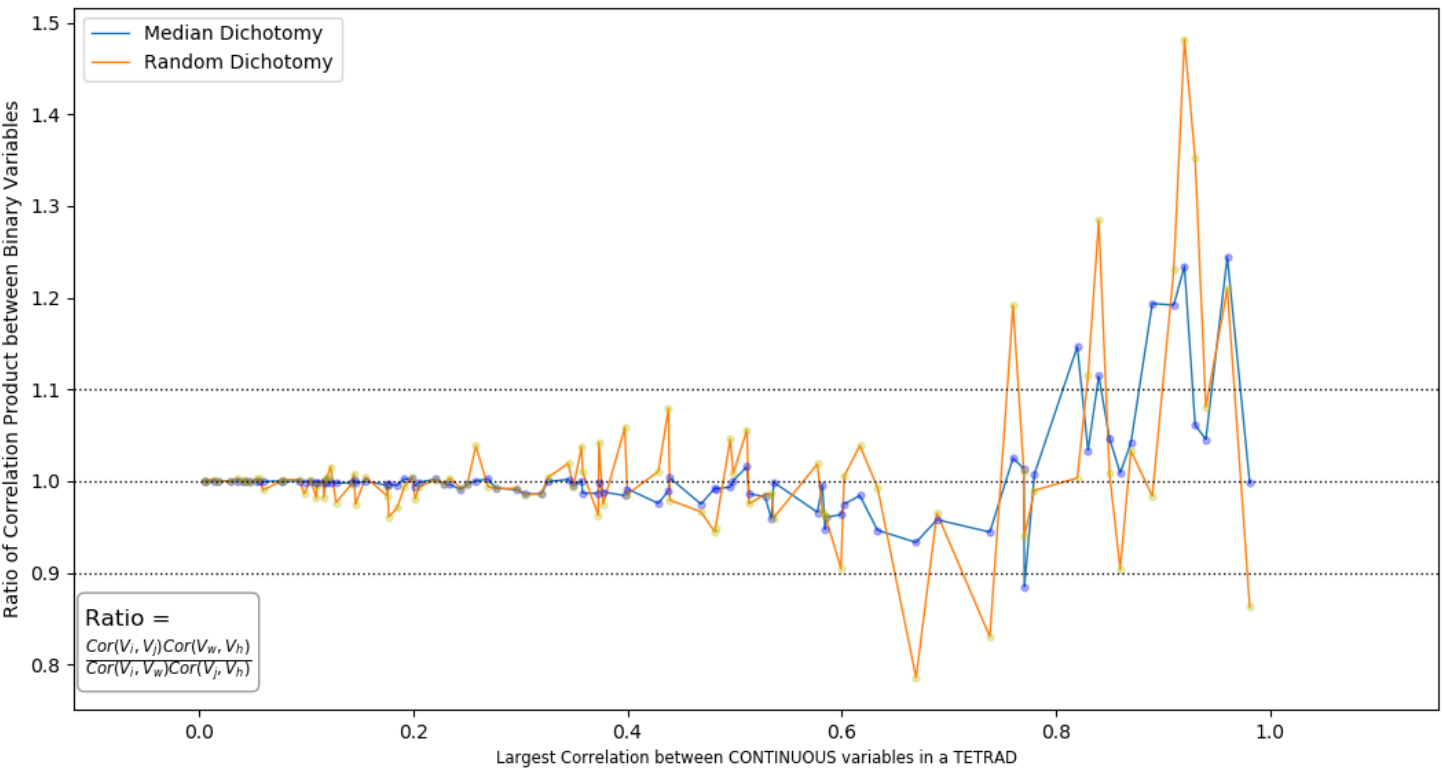}
    \caption{Ratio vs.  Largest Continuous Correlation}
    \label{fig:binaryRation}
\end{figure*}

 Figure \ref{fig:binaryRation} shows how well binary variables derived from normal Gaussian with various dichotomies satisfy the tetrad constraints: in this figure, four binary variables whose original Gaussian variables form a \model are randomly partitioned twice; the y axis is the ratio of the product of correlation between these two partitions and x axis the largest absolute value the correlation the pairs of Gaussian variables that generates the binary variables in the partition can have.  For the sake of convenience, all correlations here are positive since what matters is not being positive or negative but the absolute value. A ratio of 1 is when tetrad constraint holds. We can see that binary variables with median dichotomy satisfy the tetrad constraints with slight variations when the (absolute value of) the correlation between the standard Gaussian variables is less than 0.8; the product ratio of binary variables with non-median dichotomy fluctuates around ``1" and become less and less stable as the correlation between the standard Gaussian variables increases, suggesting a susceptibility to external noises and complex connections between mediate Gaussian variables even if each of them forms a \model .  
 
 The figure \ref{fig:binaryRation} is designed to plot Ratio versus the Largest Absolute Value of Continuous Correlation because, besides the absolute value of cutoff, (the absolute value of) continuous correlation is the only other thing that influences how well binary variables satisfy tetrad constraints.  As mentioned in setion 3.3.1, the covariance between two binary variables is closer to proportional to the correlation of the two Gaussians that generate them when the Gaussian correlation is small.  The closer to proportional the binary covariance is to the Gaussian correlation, the closer the \textit{binary tetrad}  containing the pair of binary variables satisfies the tetrad constraint (so the ratio is closer to 1) if other conditions are the same. In other words, after two partitions, the largest absolute value among the four continuous correlations determines how close the ratio is to `1'.

\subsection{Continuous-Discrete}

\begin{figure}[h]
    \centering
\begin{tikzpicture}
\begin{axis}[
  no markers, 
  domain=0:6, 
  samples=100,
  ymin=0,
  axis lines*=left, 
  xlabel=\textit{cutoff},
  every axis y label/.style={at=(current axis.above origin),anchor=south},
  every axis x label/.style={at=(current axis.right of origin),anchor=west},
  height=5cm, 
  width=12cm,
  xtick=\empty, 
  ytick=\empty,
  enlargelimits=false, 
  clip=false, 
  axis on top,
  grid = major,
  hide y axis
  ]

 \addplot [very thick,cyan!50!black] {gauss(x, 3, 1)};

\pgfmathsetmacro\valueA{gauss(1,3,1)}
\pgfmathsetmacro\valueB{gauss(2,3,1)}
\pgfmathsetmacro\valueC{gauss(3.5,3,1)}
\pgfmathsetmacro\valueD{gauss(5,3,1)}

\draw [gray]  (axis cs:1,0) -- (axis cs:1,\valueA);
\draw[gray](axis cs:2,0) -- (axis cs:2,\valueB)
    (axis cs:3.5,0) -- (axis cs:3.5,\valueC);

\draw [{Latex[ width=2mm]}-] (axis cs:0.5,0) -- (axis cs:0.5,\valueB-0.15)
    (axis cs:0.5,0) -- (axis cs:0.5,\valueB-0.15);

\node[below] at (axis cs:0.5, \valueB-0.07)  {$\textbf{0}$}; 
\node[below] at (axis cs:1, 0)  {$S_{i0}$}; 
\node[below] at (axis cs:1.5, \valueA+0.04)  {$\textbf{1}$}; 
\node[below] at (axis cs:2, 0)  {$S_{i1}$}; 
\node[below] at (axis cs:2.7, 0)  {......}; 
\node[below] at (axis cs:3.5, 0)  {$S_{in-1}$};
\node[below] at (axis cs:4.3,  \valueA+0.04)  {$\textbf{n}$};

\end{axis}
\end{tikzpicture}
\caption{Linear Discretizaion}
\label{fig:LinearDisc}
\end{figure}
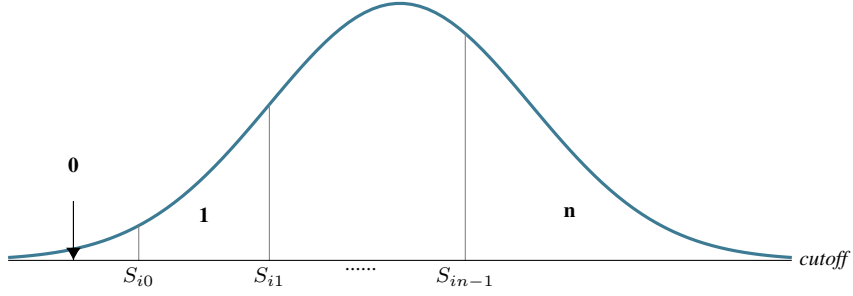

FOFC works effectively on discrete variables with randomly many categories as long as each of them is discretized from a Gaussian variable in the same way the binary variables are generated in the last section.  Roughly speaking, as the number of categories increases the performance of FOFC increases.  As will be discussed later in this section, the reason behind such performance varies with the number of possible values the variables can take.  All the variables discussed in this section are derived from standard Gaussian variables with the same discretization mechanism shown in the Figure \ref{fig:LinearDisc}:

\textit{for a discrete variable $V_i$, there is a standard Gaussian variable $X_i$ such that $V_i=n$ \textbf{iff} $S_{i(n-1)}<X_i\leq S_{in}$}.

\pgfmathdeclarefunction{gauss}{3}{%
  \pgfmathparse{1/(#3*sqrt(2*pi))*exp(-((#1-#2)^2)/(2*#3^2))}%
}

\begin{remark*}
For any two discrete variables $V_i$ and $V_j$ that are derived from standard Gaussian variable $X_i$ and $X_j$ with linear discretization, such that $V_i$ has k possible values and $V_j$ has g possible values, their covariance $Cov(V_i, V_j)$ is:
\begin{center}

     $\sum\limits_{c_i=0}^{k-2}\sum\limits_{c_j=0}^{g-2}\Psi(S_{ic_i},S_{jc_j}, \rho_{ij})-\Psi(S_{ic_i},S_{jc_j}, 0)$
\end{center}
where $\rho_{ij}$ is the correlation between $X_i$ and $X_j$ and $S_{ic_i}$ denotes the cutoff value of $V_i$ such that $V_i=c_i$ when $S_{i(c_i-1)}<X_i\leq S_{ic_i}$.
\end{remark*}
This remark can be easily seen by a small proof by induction in the appendix.

According to the \textbf{remark}, if the absolute value of every cutoff of each variable is small enough (smaller than 1), using the same idea of \textbf{section 2.3.3 Non-median Dichotomy}, the covariance between any two discrete variables with $k$ and $g$ possible values can be approximated:
\begin{align} 
\begin{split}
& Cov(V_i, V_j)\ \\
&= \sum\limits_{c_i=0}^{k-2}\sum\limits_{c_j=0}^{g-2}\Psi(S_{ic_i},S_{jc_j}, \rho_{ij})-\Psi(S_{ic_i},S_{jc_j}, 0)\ \\
&\approx \int_0^{\rho_{ij}}\dfrac{1}{2\pi\sqrt{1-r^2}}[ \sum\limits_{c_i=0}^{k-2}\sum\limits_{c_j=0}^{g-2}exp\{\frac{S_{ic_i}^2+S_{jc_j}^2}{-2}\}]dr\ \\
&= \sum\limits_{c_i=0}^{k-2}exp\{\frac{S_{ic_i}^2}{-2}\}\sum\limits_{c_j=0}^{g-2}exp\{\frac{S_{jc_j}^2}{-2}\}\int_0^{\rho_{ij}}\dfrac{1}{2\pi\sqrt{1-r^2}}dr\ \\
&=\sum\limits_{c_i=0}^{k-2}exp\{\frac{S_{ic_i}^2}{-2}\}\sum\limits_{c_j=0}^{g-2}exp\{\frac{S_{jc_j}^2}{-2}\}\dfrac{1}{2\pi}arcsin(\rho_{ij})\ \\
\end{split}
\end{align}
Recall that $\rho_{ij}$ is the covariance between $X_i$ and $X_j$.  As mentioned in \textbf{section 2.3.3}, $arcsin(\rho_{ij})\approx\rho_{ij}$ when $|\rho_{ij}|$ is not too large.  To see whether the tetrat constraints approximately hold between discrete variables generated through linear discretization from continuous variables  in a \model when their correlations are not too large, let us check the tetrads between discrete variables in figure \ref{fig:clusterCD} with arbitrary number of categories. More specifically, variable $V_i$ has $k_i$ many categories:
\begin{align*} 
\begin{split}
& Cov(V_1, V_2)\ \\
&\approx \sum\limits_{c_1=0}^{k_1-2}exp\{\frac{S_{1c_1}^2}{-2}\}\sum\limits_{c_2=0}^{k_2-2}exp\{\frac{S_{2c_2}^2}{-2}\}\dfrac{1}{2\pi}arcsin(\rho_{12})\ \\
&\approx \sum\limits_{c_1=0}^{k_1-2}exp\{\frac{S_{1c_1}^2}{-2}\}\sum\limits_{c_2=0}^{k_2-2}exp\{\frac{S_{2c_2}^2}{-2}\}\dfrac{1}{2\pi}\rho_{12}\ \\
&\approx \sum\limits_{c_1=0}^{k_1-2}exp\{\frac{S_{1c_1}^2}{-2}\}\sum\limits_{c_2=0}^{k_2-2}exp\{\frac{S_{2c_2}^2}{-2}\}\dfrac{1}{2\pi}a_1a_2\ \\
\end{split}
\end{align*}
Similarly, 
\begin{align*}
    \begin{split}
         &Cov(V_3,V_4)\approx 
         \sum\limits_{c_3=0}^{k_3-2}exp\{\frac{S_{3c_3}^2}{-2}\}\sum\limits_{c_4=0}^{k_4-2}exp\{\frac{S_{4c_4}^2}{-2}\}\dfrac{1}{2\pi}a_3a_4\ \\
\end{split}
\end{align*}
And it's easy to see:
\begin{center}
    $Cov(V_1, V_2)Cov(V_3,V_4)\approx Cov(V_1, V_3)Cov(V_2,V_4)$
\end{center}

So the tetrad constraints approximately hold between discrete variables generated through linear discretization from continuous variables in a \model when their correlations are not too large.
\subsubsection{Discrete Variables with Large Number of Categories Behave similarly to the Continuous}
As shown above, the tetrad constraints approximately hold between discrete variables generated from continuous variables in a \model.  The satisfaction of the tetrad constraints requires that correlations between the continuous variables are not too large and also the absolute values of cutoffs ($|S_{ic_i}|$) are small (to enable the approximation in \textbf{2.6} and \textbf{2.9}). 

\begin{figure}
    \centering
    \includegraphics[width=\linewidth]{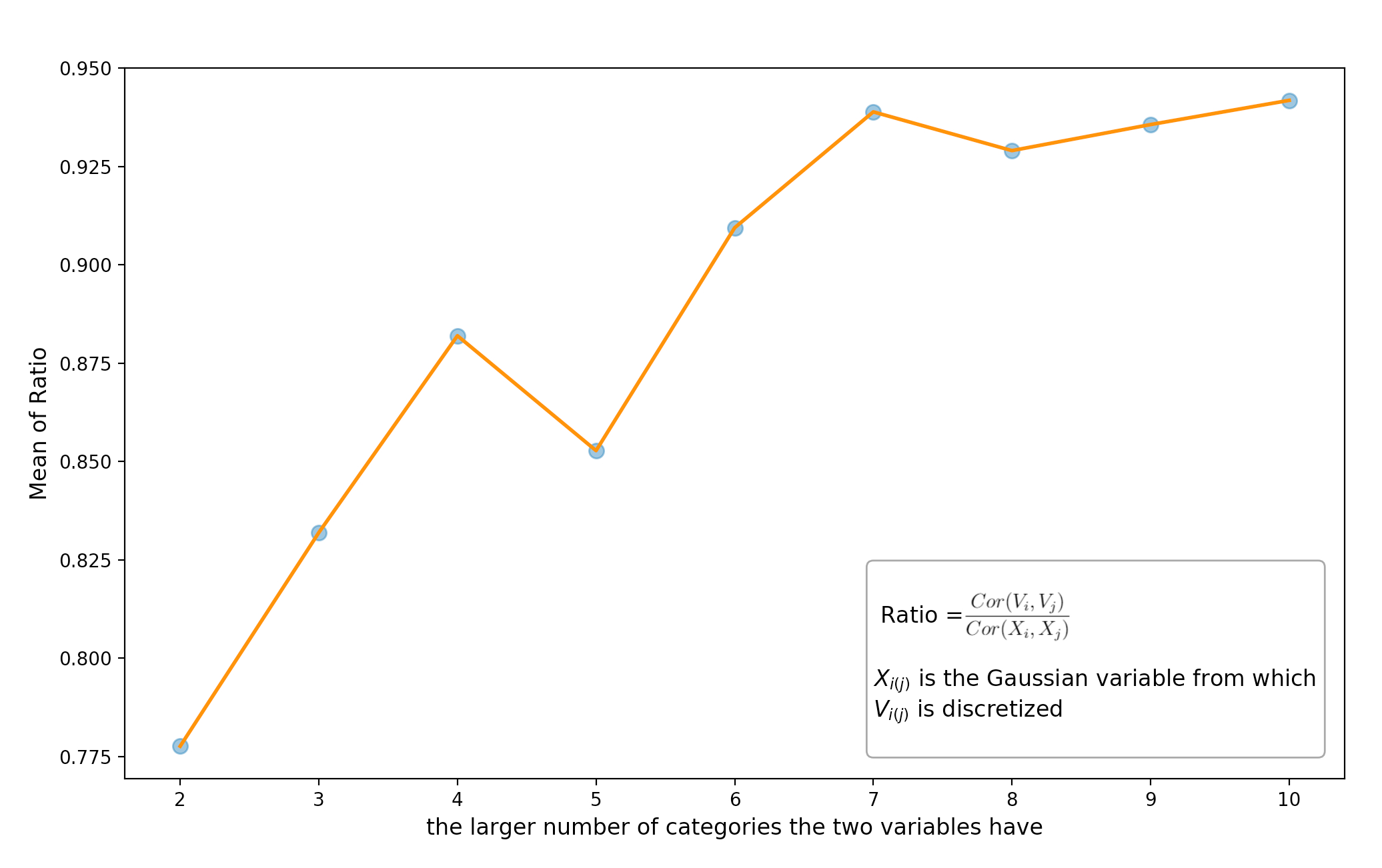}
    \caption{Ratio of Correlation vs Number of Categories}
    \label{fig:numCatratio}
\end{figure}

On the other hand, the more categories a standard Gaussian variable is discretized into, the more similarly the resulting discrete variable behaves to the original Gaussian. Figure \ref{fig:numCatratio}, which is a plot of the mean of ratio of the correlation between two discrete variables over the correlation between the two original Gaussian variables versus the number of categories a discrete variable takes, illustrates this feature.

The figure is generated by first calculating the \textbf{mean of ratio}:
A 9x2 Matrix \textit{M} is initialized to set the number of categories each pair of discrete variables are taking. $M$ represents 9 pairs of variables; one of the two variables in the \textit{i}th pair takes $M[i,0]$ possible values and another  $M[i,1]$ .  The value of each entry of $M$ is set up like this: $M[i,1] = i+2$ $(i\in \{0...8\}$ ) and $M[i,0]$ is a random number between $2$ and $M[i,1]$.  Let $i_m$ denote the variable that has more categories in $i$th pair and $\#(i_m)$ the number of categories $i_m$ takes.  The setup of $M$ entails that $\#(i_m)>\#(j_m)$ if $i>j$.

For a pair of discrete variables of which the numbers of possible values are $w$ and $z$, we calculate the correlation of the pair given the correlation between the Gaussian variables, as an element in \textit{ContCor}, from which they are derived.  Two cutoff arrays, one of length $w-1$ and another of $z-1$, were initialized such that each array has an increasing order and every two adjacent elements differ from a random number between 0 and 1.  The two cutoff arrays are then centered to get a zero mean\footnote{the centering operation is just to make sure that some cutoffs are negative and some are positive; there is no difference in the final result depending on whether the cutoff array is centered or not}.  Given the correlation between the two standard Gaussian variables, the variance of each discrete variable is first calculated, then their covariance using the formula in \textbf{remark}, then their correlation, which is then divided by the Gaussian correlation to get the ratio.  For every pair of discrete variables, such a ratio for every element in \textit{ContCor} is calculated, then all the ratios are summed up and divided by the length of \textit{ContCor} to get the \textbf{mean of ratio}.  After collecting the \textbf{mean of ratio} for every pair of variable, the figure is plotted such that the x axis is the second column of $M$ and for a point with $M[j,1]$ as the x-coordinate, its y-coordinate is the \textbf{mean of ratio} for the \textit{i}th pair.

As shown in this figure, the correlation between the discrete variables approaches the original Gaussian correlation closer and closer as the number of category increases.  In fact, when the larger number of category exceeds six, the difference between the discrete correlation and continuous correlation is less than 10\%.

\section{Test Results}
This section presents the performance of FOFC on simulation data: the first part of the performance presented here is about FOFC running on data with various types and comparison of various correlations: sample variance, tetrachoric correlation and rank correlation; the second part is a comparison of performance between FOFC and a clustering algorithm working on data with mixed types called MGM\cite{MGM}.

\subsection{Performance on Continuous and Discrete data}
\subsubsection{General Setup of Simulation}
\begin{figure}
    \centering
    \includegraphics[width=\linewidth]{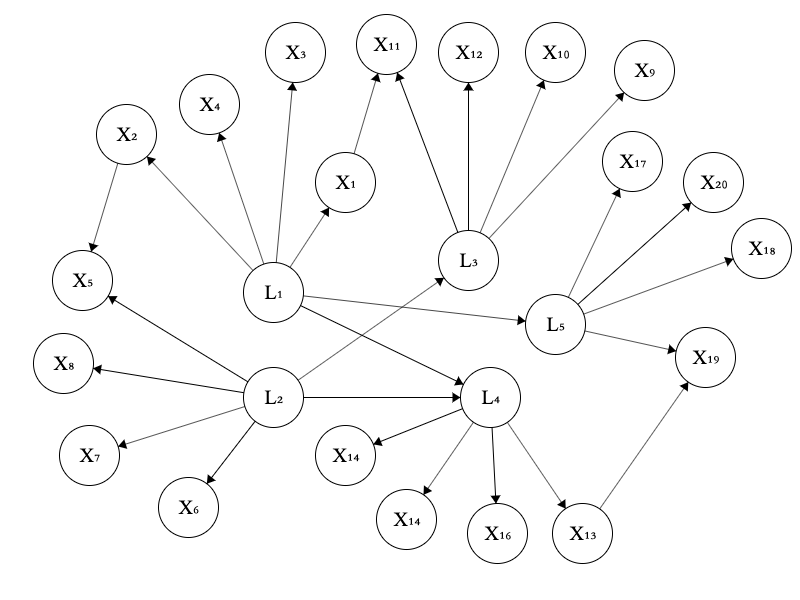}
    \caption{A Graph used to simulate Gaussian Data with 3 impurities}
    \label{fig:L5E4imp}
\end{figure}

\begin{figure}
    \centering
    \includegraphics[width=\linewidth]{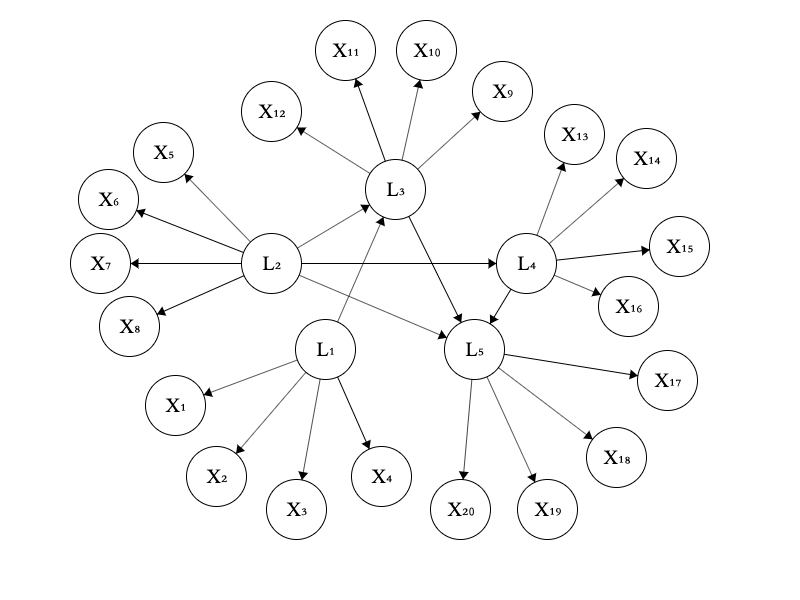}
    \caption{A Graph used to simulate Gaussian Data}
    \label{fig:L5E6}
\end{figure}

This section presents the performance of FOFC on Gaussian data, binary data and data with more than two possible values discretized from the original continuous variables. The original continuous data is simulated from a graph with five latent variables each having four measured children.  In the simulation study, the number of edges connecting the latents can be zero, one, three, six, seven or nine.  Figure \ref{fig:L5E6} shows one example where the graph has 5 \modelp with 6 edges connecting the five latents.  These numbers are chosen to represent cases from those in which all latents are independent from each other to those in which the latents are densely connected.
Note that the ability of the algorithm to detect impurities of the model, i.e., one measured variable is directly caused by more than one latent variables or measured variables directly causes other measured variables, is tested but is not explicitly shown, because such ability is found to be stable regardless of the data type.  The graphs used for the simulation contains impurities, such as figure \ref{fig:L5E4imp}, where there are three pairs of measured variables having direct causal relations.

After the Gaussian data is simulated, if asked to discretize, the system will generate random cutoffs for each variable to linearly discretize the variable into the number of categories set up beforehand.  The discrete data shown in this section belong to the following situations, each situation under three sample sizes(100, 500 and 2000):
\begin{itemize}
    \item binary variable from median dichotomy
    \item binary variable from non-median dichotomy
    \item data with THREE possible values
    \item data with FOUR possible values
    \item data with FIVE possible values
    \item data with SIX possible values
    \item data with EIGHT possible values
\end{itemize}
For the sake of convenience, the simulation is set up so that, when running on the discretized data, all variables included have the same number of possible values.  This should not raise a concern because as proven in \textbf{section 3.4},  whether the measured discrete variables take the same number of possible values or not does not influence whether they satisfy the tetrad constraints.

\subsubsection{Precision and Recall}
The two indexes used to evaluate the performance of the algorithm are the very common ones: \textit{precision} and \textit{recall}. The performance for each index is shown by a figure. 

One expects FOFC to identify latent variables through measured ones and ultimately recover the connection between the latents. \textit{Precision} measures that, for each group of variables clustered by the algorithm, how accurate it is for them to actually form a \model:  for an estimated cluster, the system will first find the true cluster that contains the largest number of the variables in this estimated cluster and calculate the percentage of variables in the estimated cluster contained in the true cluster as an \textit{individual precision} for this single estimated cluster.  In the best case where all the variables in an estimated cluster indeed forms the same \model, the \textit{individual precision} will be one; in the worst case the \textit{individual precision} will be $1/n$ where $n$ is the number of variables in the estimated cluster. The system calculates the \textit{precision} for each estimated cluster, then calculate the mean of all \textit{individual precisions} as the \textit{final precision} for the simulation.  

\begin{figure}
    \centering
    \includegraphics[width=\linewidth]{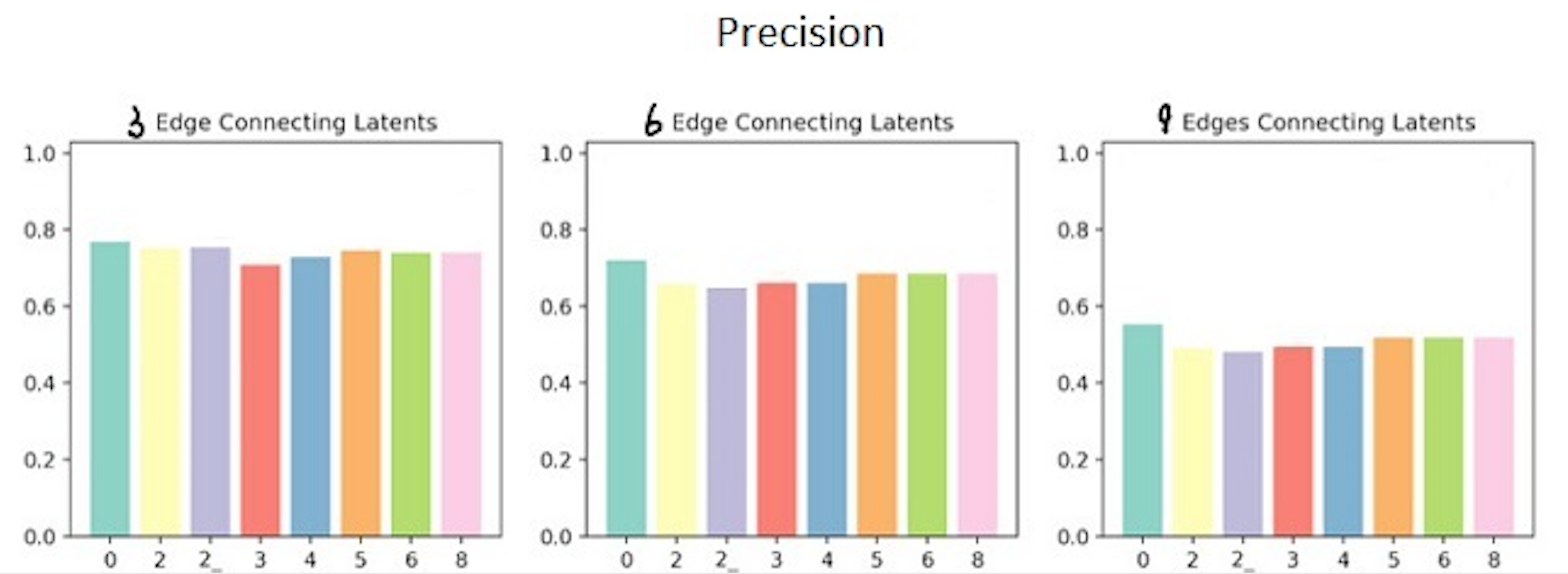}
    \caption{Precision of FOFC with the sample size of 100}
    \label{fig:precision100}
\end{figure}

\begin{figure}
    \centering
    \includegraphics[width=\linewidth]{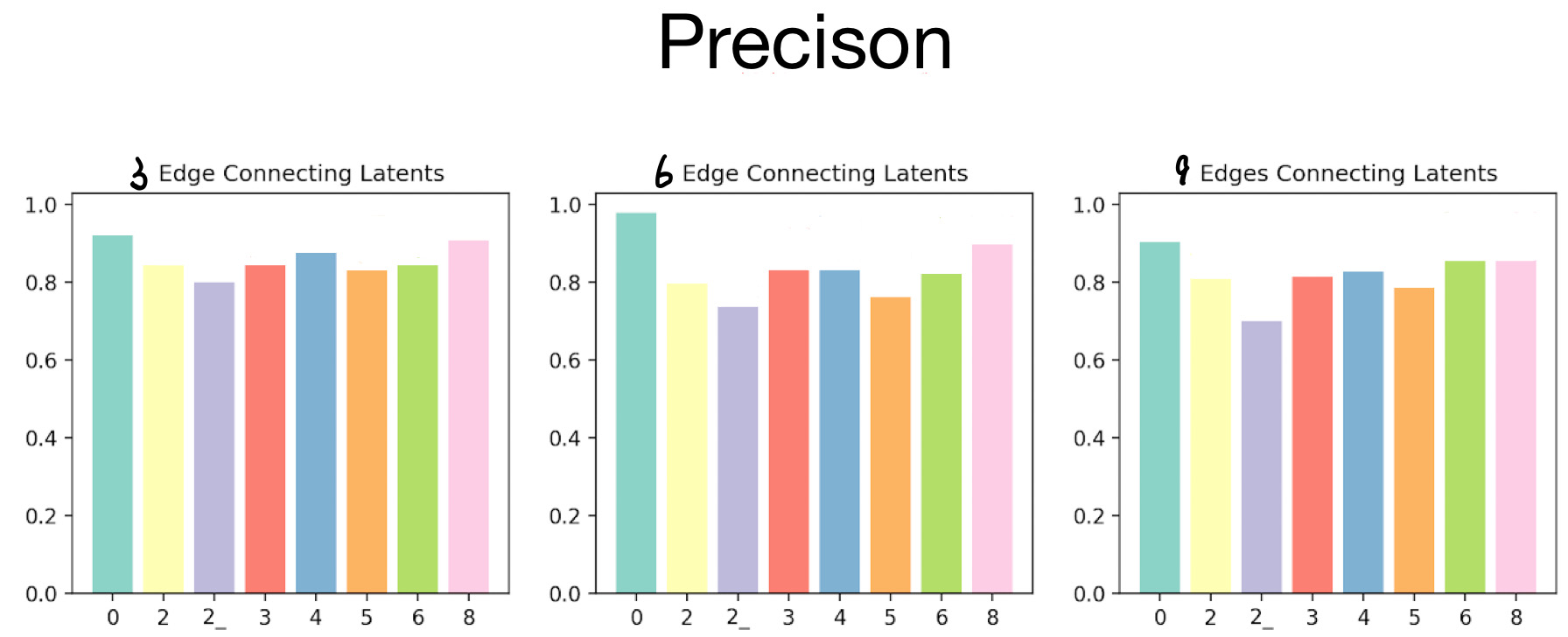}
    \caption{Precision of FOFC with the sample size of 500}
    \label{fig:precision500}
\end{figure}

\begin{figure}
    \centering
    \includegraphics[width=\linewidth]{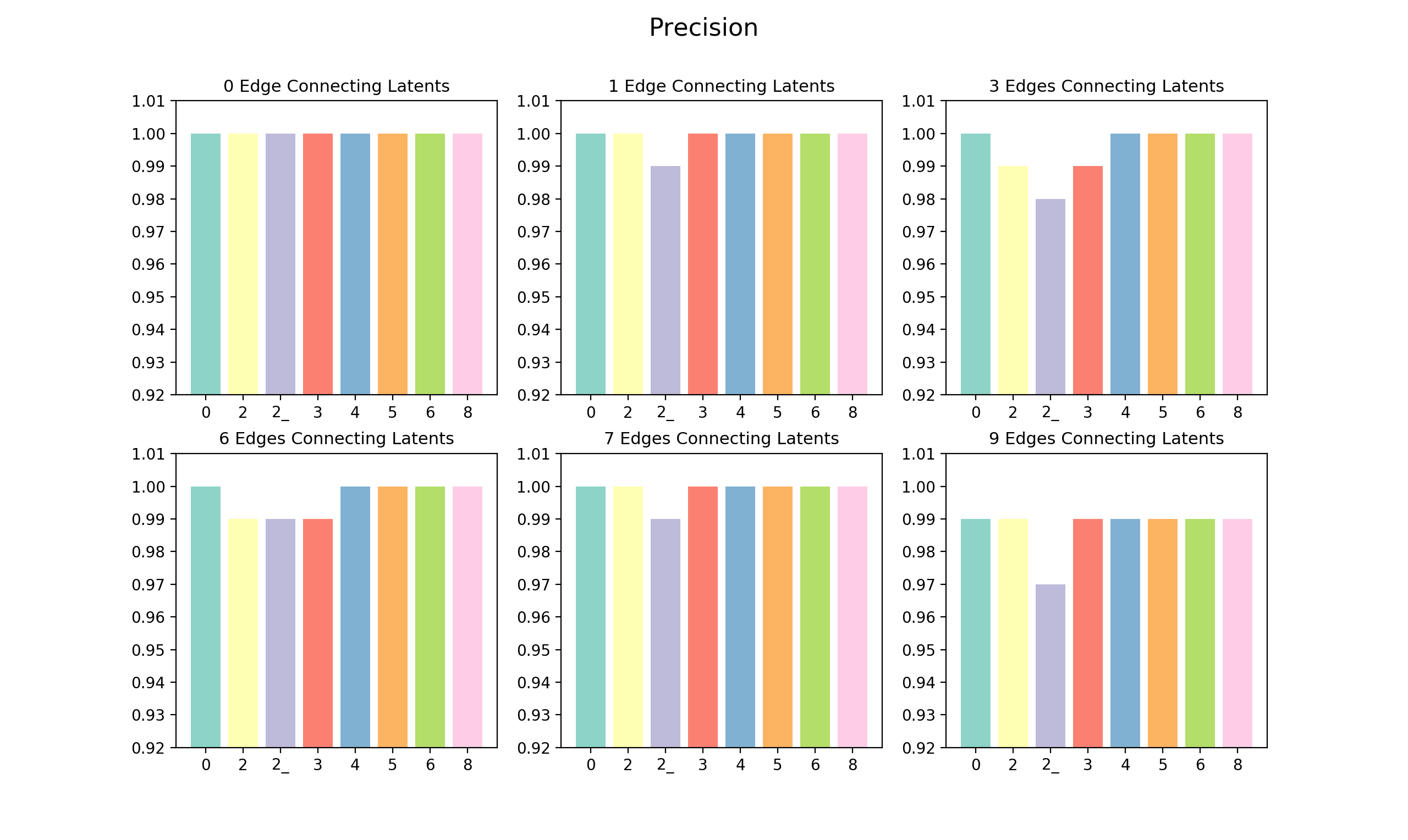}
    \caption{Precision of FOFC with the sample size of 2000}
    \label{fig:precision}
\end{figure}

Figure \ref{fig:precision100},\ref{fig:precision500} and \ref{fig:precision} shows the precision of FOFC performed on datasets with different sizes.  The dataset with sample size 100 and 500 is simulated from graphs where there are 3, 6, or 9 edges connecting the five latent variables(such as figure \ref{fig:L5E6}). With sample size 2000, FOFC is tested on simulations where the true graph is in six different conditions.  As mentioned before, the difference between the six conditions is the density of the connection between the five latent variables.  The connection is introduced as the title of each of the six sub-figures.  For each sub-figure, the x-axis is the type of data the algorithm runs on: 0 means the data is Gaussian; 2 indicates binary data with median dichotomy; $2\_$ binary data with non-median dichotomy; 3 data with three possible values; 4 data with four possible values, etc; the y-axis is the mean of \textit{final precision} of the 40 simulations.  Note that the 40 simulations are based on the same true graph.

\begin{figure}
    \centering
    \includegraphics[width=\linewidth]{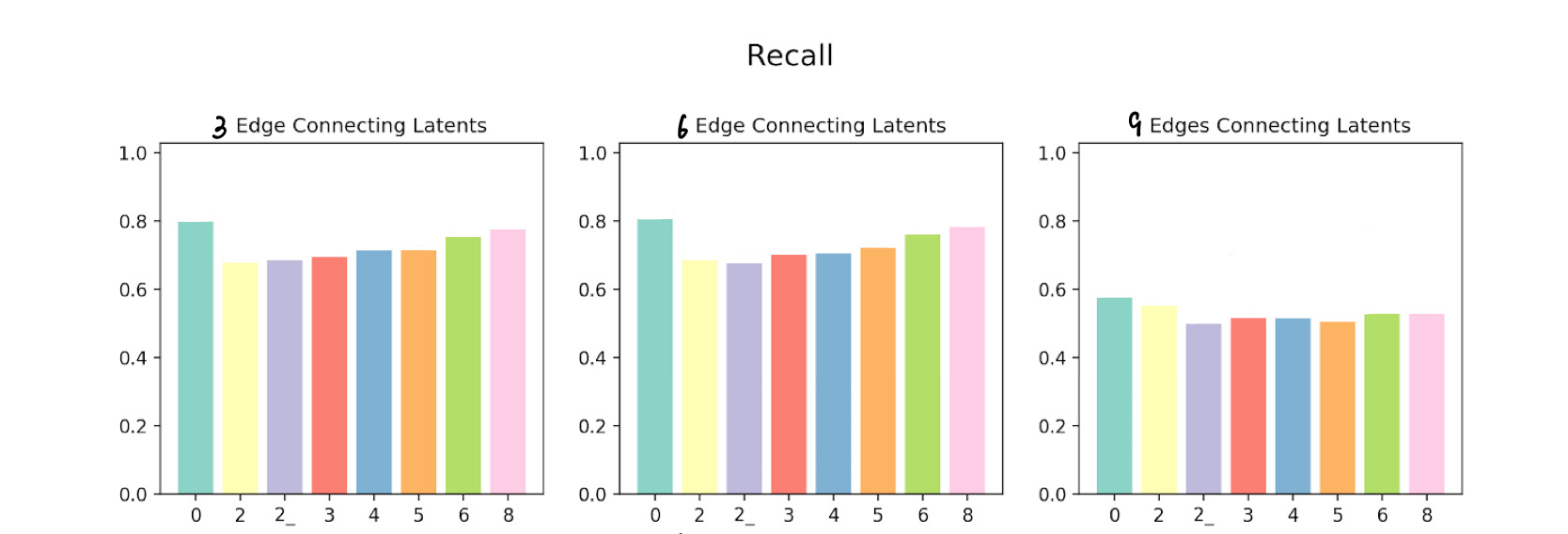}
    \caption{Recall of FOFC with the sample size of 100}
    \label{fig:recall100}
\end{figure}

\begin{figure}
    \centering
    \includegraphics[width=\linewidth]{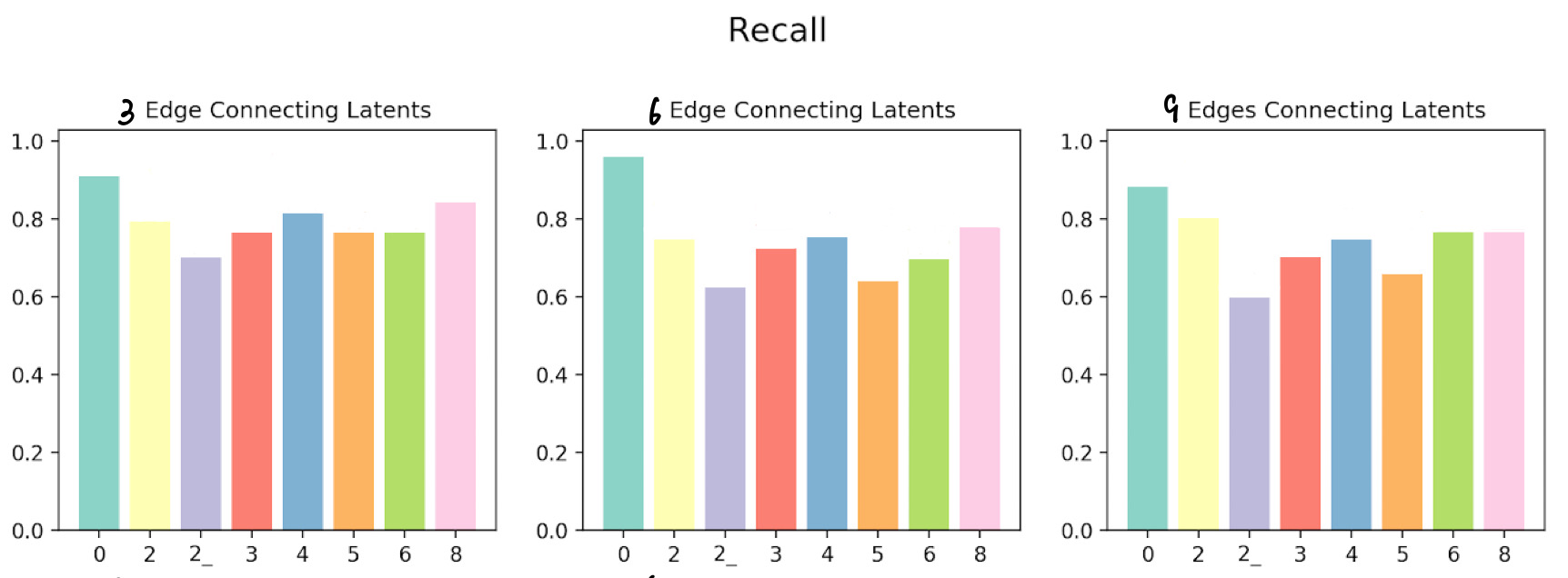}
    \caption{Recall of FOFC with the sample size of 500}
    \label{fig:recall500}
\end{figure}

The precision of FOFC increases as the sample size increases.  The reliably high precision in the figure \ref{fig:precision} indicates that, regardless of the situation of the data, if the algorithm detects a latent common cause from a group of variables by clustering them together, it is safe to say that these variables indeed share a latent common cause, forming a \model.

\textit{Precision} measures the accuracy of FOFC on different data types whereas \textit{recall} measures the sensitivity, i.e., for each \model (or true cluster) in the true graph, how likely it is for the algorithm to actually put all the variables it contains into an estimated cluster:  for an true cluster, the system will first find the estimated cluster that contains the largest number of the variables in this true cluster and calculate the percentage of variables in the true cluster contained in the estimated cluster as an \textit{individual recall} for this single true cluster.  In the best case where all the variables in an true cluster indeed are grouped into the same estimated cluster, the \textit{individual recall} will be one; in the worst case the \textit{individual recall} will be 0 when all variables in the true cluster are determined as independent from others. The system calculates the \textit{recall} for each true cluster, then calculate the mean of all \textit{individual recalls} as the \textit{final recall} for the simulation.  

\begin{figure}
    \centering
    \includegraphics[width=\linewidth]{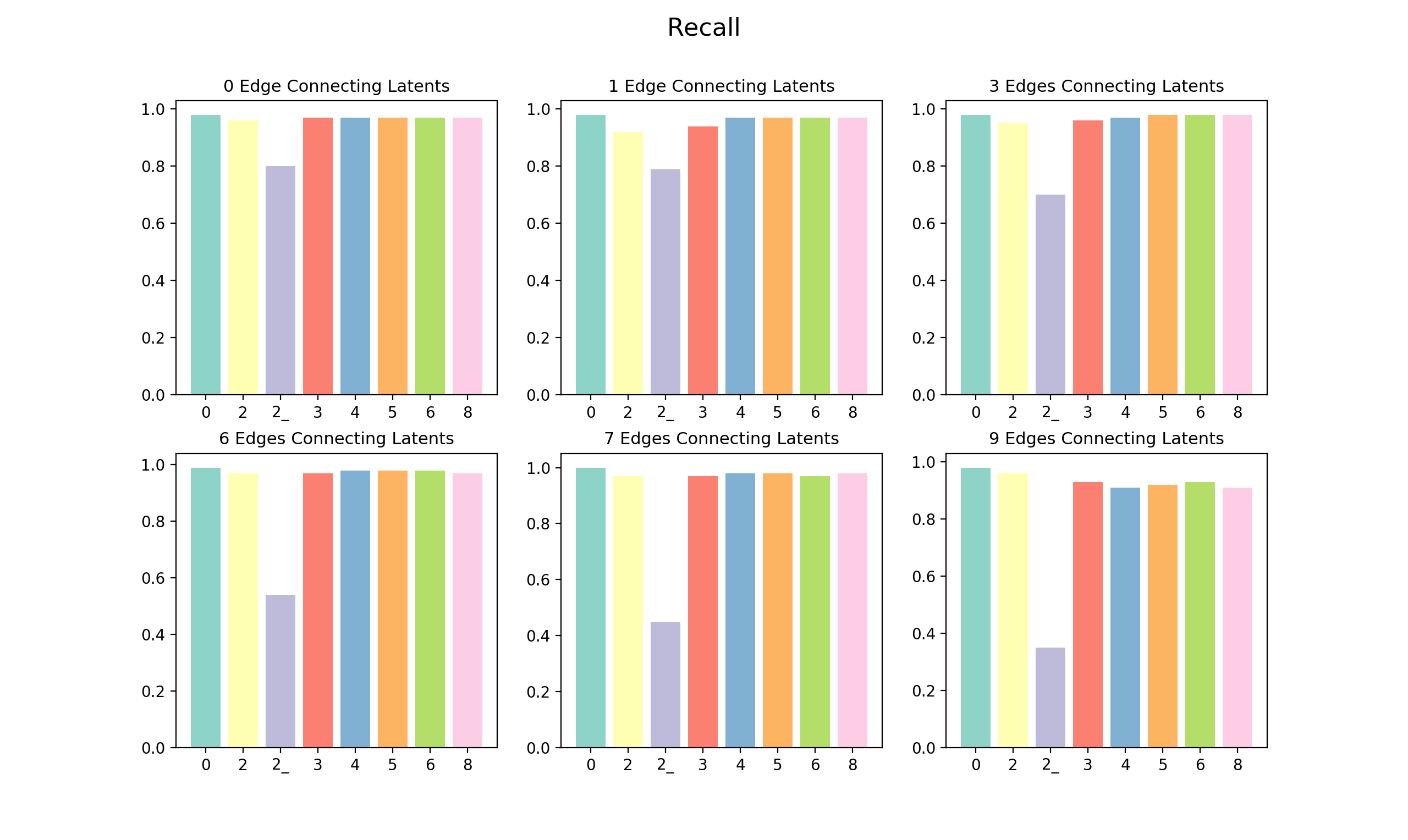}
    \caption{Recall of FOFC with the sample size of 2000}
    \label{fig:recall}
\end{figure}

Figure \ref{fig:recall100}, \ref{fig:recall500} and \ref{fig:recall} show the recall of the FOFC tested on the datasets from which the \textit{precisions} with the corresponding sample sizes shown above are calculated.  The connection between the five latent variables is introduced as the title of each of the six sub-figures.  For each sub-figure, the x-axis has the same meaning as before; the y-axis is the mean of \textit{final recall} from the 40 simulations each time.

\begin{figure}
    \centering
    \includegraphics[width=\linewidth]{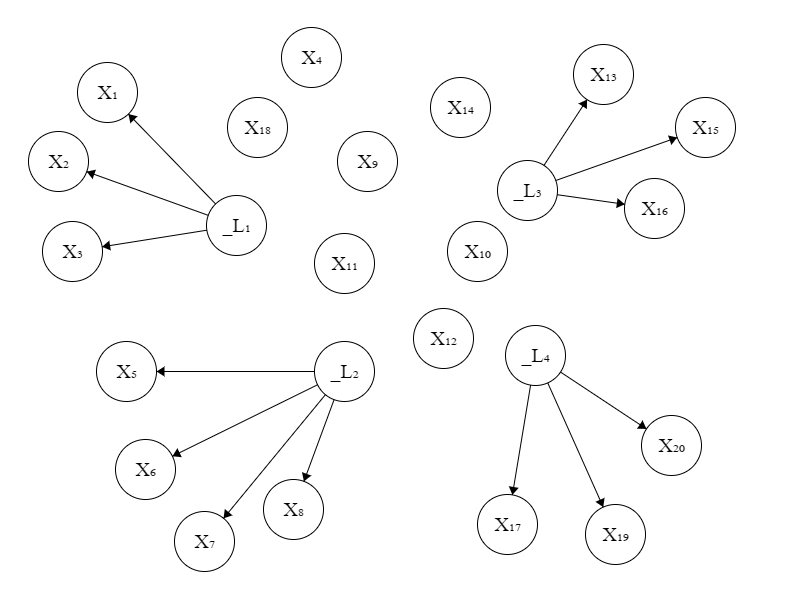}
    \caption{Clustering Result on Binary data with Non-Median Dichotomy}
    \label{fig:L5E6NM}
\end{figure}

The \textit{recall} of the FOFC running on each data type is similar to \textit{precision} except for the binary data with non-median dichotomy.  As the number of edges connecting the five latent variables increases, the \textit{recall} for non-median binary data decreases from 0.8 to less than 0.4, showing that the algorithm gradually loses the ability to identify pure clusters using the tetrad constraints.  The high \textit{precision} and low \textit{recall} indicate that, when the data is non-median binary, FOFC can form correct clusters but can only recover some of them while falsely leaves many variables outside of clusters which they form in the true graph.  A typical output of the algorithm in this situation is shown in figure \ref{fig:L5E6NM} when the true graph is figure \ref{fig:L5E6}.  Since the algorithm is run on binary data instead of the original continuous data, all the $X_i$ variables in figure \ref{fig:L5E6} represents the Gaussian variable and in figure \ref{fig:L5E6NM} the binary variable discretized from the original. $\_L_i$ in figure \ref{fig:L5E6NM} represents the latent variable the algorithm postulated for each estimated cluster.  Another thing worth mentioning is that, although FOFC is designed to study the latent variables, the algorithm itself does not calculate the connections between the latent variables and will not put edges between them.

 As pointed in \textbf{section 3.3.2}, when the binary data is discretized with a random number between zero and one as the cutoff value whereas the original Gaussian variable has a mean of zero, although the correlation between the binary variables of which the parents form a \model approximately satisfy the tetrad constraints, the clustering decision is disturbed by the connections between the latent variables.  The disturbance becomes inevitable for at least two reasons.  First, the absolute value of the correlation shrinks after the dichotomy, making the algorithm more likely to judge two variable as independent even if they are not, not to mention testing the tetrad constraints.  Second, the variables only approximately satisfy the tetrad constraints; as mentioned before, multiple factors can influence how well the constraints are satisfied, such as how large a cutoff value is.  The variables not belonging to the same cluster will have non-zero correlation when the latent variables are not independent, making it harder for the algorithm to test whether certain group of variables satisfy the tetrad constraints or not.
 
\subsection{Testing Rank Correlation and Tetrachoric Correlation}

Rank correlation and tetrachoric correlation calculate the correlation between discrete variables.  Rank correlation measures the ordinal association between two discrete variables.  The tetrachoric correlation is the specific case of polychoric correlation: given two discrete variables are generated from two Guassian variable, the polychoric correlation estimates the association between the underlying Guassian variables from the discrete variables, and tetrachoric correlation is the case where the discrete variables are binary.

This section presents the performance of FOFC when the rank and tetrachoric correlation are used in the algorithm. That is, instead of using correlations, the variations use Pearson rank correlations or tetrachoric correlations.  The setup of the true graph and how the precision and recall are calculated are the same as before.  For a graph with 3, 6 or 9 edges connecting the five latent variables, 40 simulations are generated with 500 data points each time.  

\begin{figure}
    \centering
    \includegraphics[width=\linewidth]{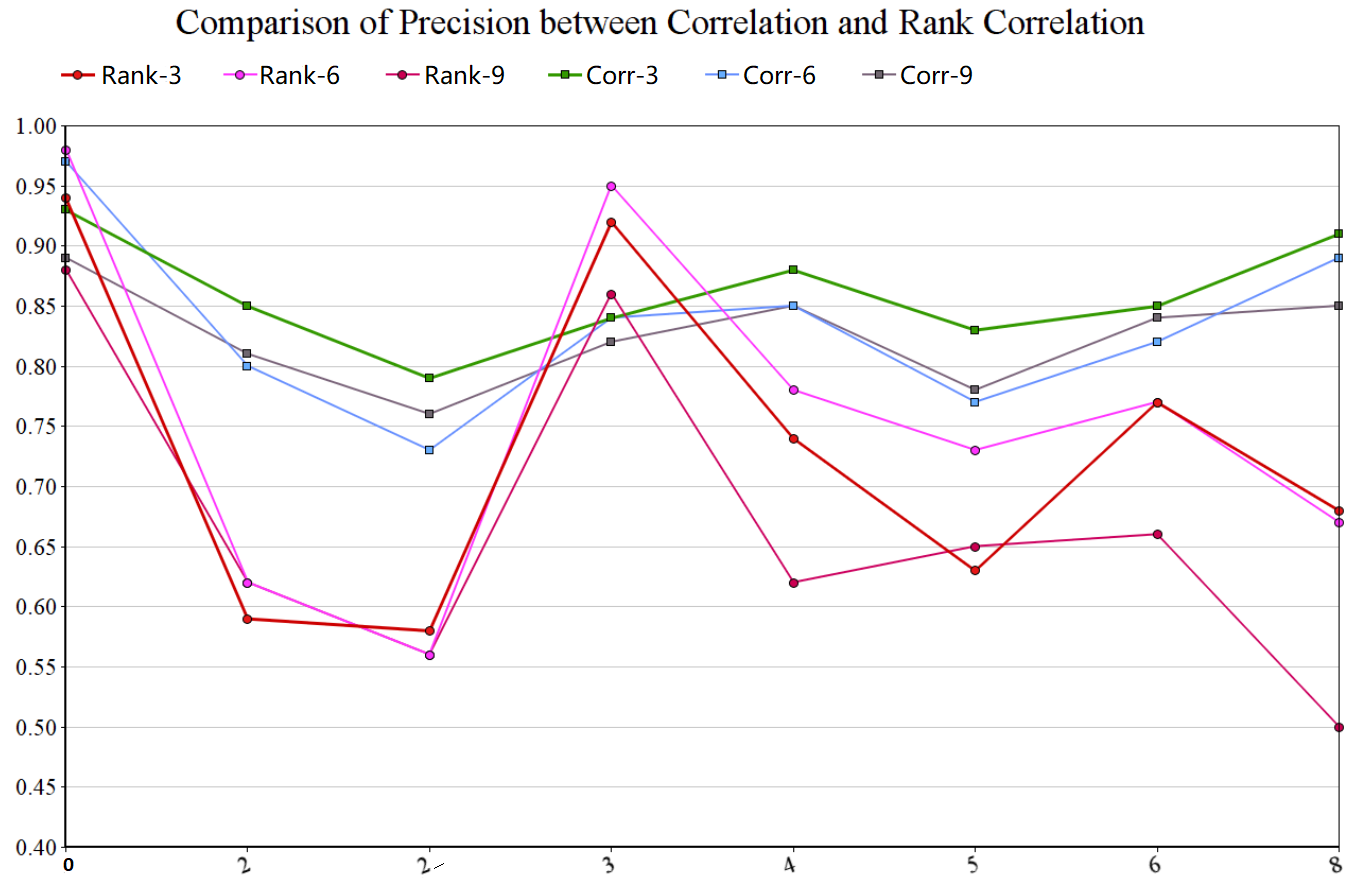}
    \caption{Precision of Discretized variable with Rank Correlation vs Correlation}
    \label{fig:rankPrecis}
\end{figure}

\begin{figure}
    \centering
    \includegraphics[width=\linewidth]{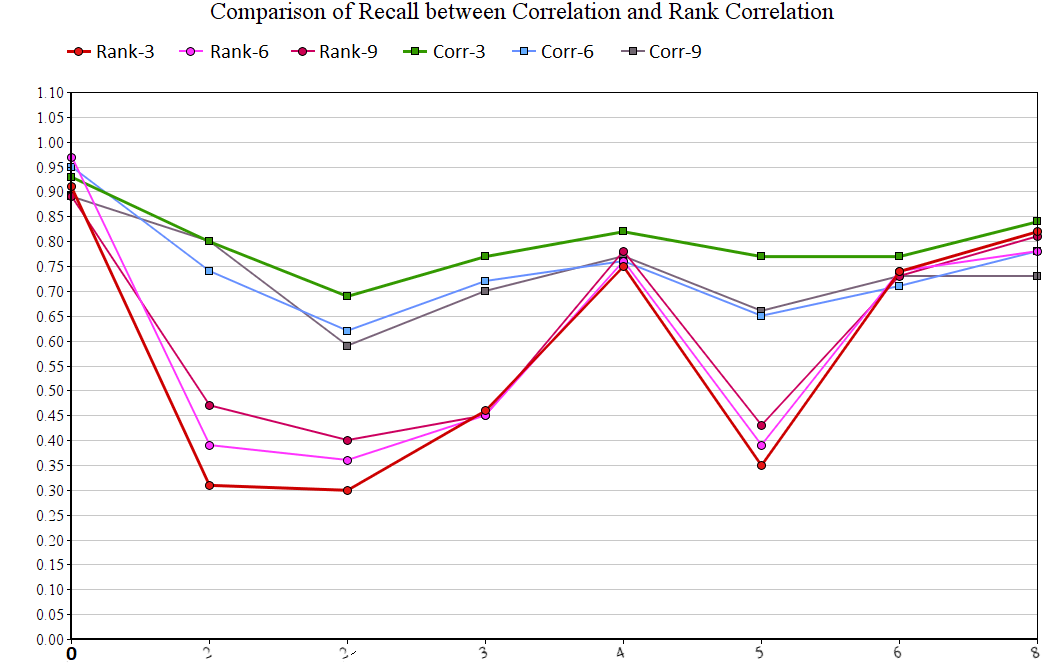}
    \caption{Recall of Discretized variable with Rank Correlation vs Correlation}
    \label{fig:rankRecall}
\end{figure}

Figure \ref{fig:rankRecall} and \ref{fig:rankPrecis} compares the recall and precision of FOFC running with rank correlation with the result we get before.  As before, the x-axis denotes how many categories the variables are discretized into; the y-axis is the final mean of 40 simulations of either recall or precision, depending on the graph. The meaning of each line is explained on the top of the graph, which consists of two part: what kind of correlation is calculated and the number of edges connecting the latent variables in the graph from which the data is generated. For instance, the ``\textbf{Rank-3}" in figure \ref{fig:rankRecall} means that this line denotes the result calculating \textbf{rank correlation} when the five latent variables are connected with \textbf{3} edges in the true graph. 

Note that each graph includes the performance of FOFC on Gaussian variables (it is the statistics corresponding to the 0 on the x-axis).  Since in this case the variable is continuous, the variation is always calculated as correlation (so, even if in figure \ref{fig:rankRecall}  the red line ``Rank-3" starts with the point (0, 0.91), this 0.91 recall is for the FOFC running a continuous variables, therefore the variation used here is correlation) . Comparing to the perfomance of FOFC in the last section with the same sample size, we see that the only statistics higher in the rank correlation is the precision for the case where variables have three possible values.  In any other cases, including the recall of rank correlation for the variables taking three possible values, the algorithm performs better when calculating the correlation.  One possible explanation is that the rank correlation does not reflect linearity, which is crucial to the performance of FOFC when the variables are discretized.

\begin{figure}
    \centering
    \includegraphics[width=\linewidth]{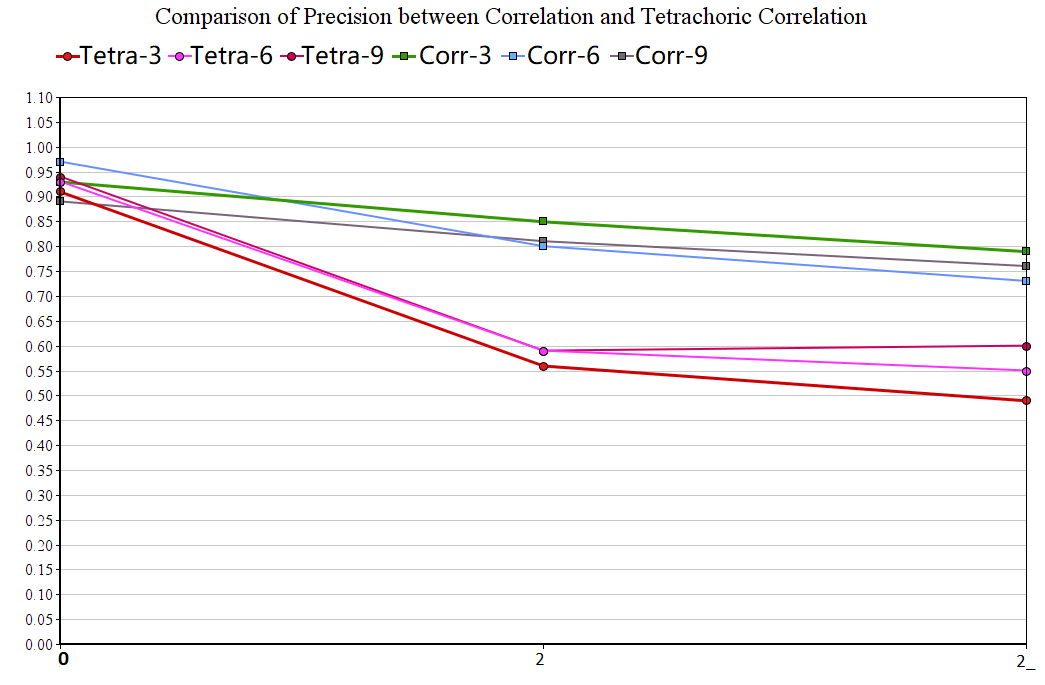}
    \caption{Precision of Discretized variable with Tetrachoric Correlation vs Correlation}
    \label{fig:PrecTetra}
\end{figure}

\begin{figure}
    \centering
    \includegraphics[width=\linewidth]{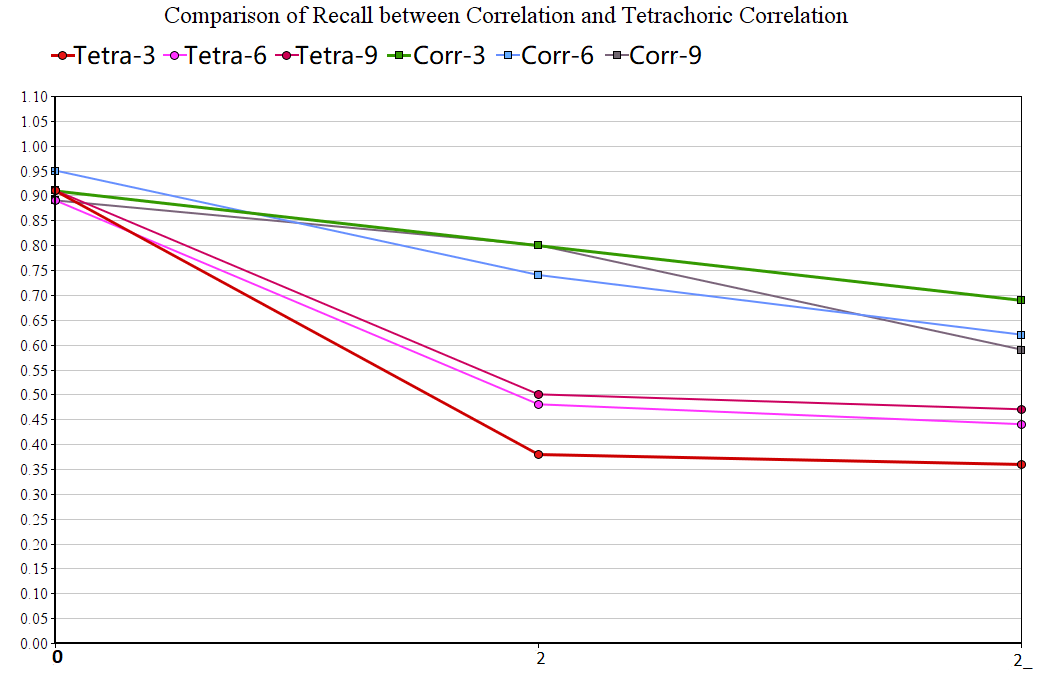}
    \caption{Recall of Discretized variable with Tetrachoric Correlation vs Correlation}
    \label{fig:recallTetra}
\end{figure}

Figure \ref{fig:recallTetra}
 and \ref{fig:PrecTetra} compares the recall and precision of FOFC running with tetrachoric correlation with the result we get before.  In each graph, the y-axis has the same meaning as before, and the x-axis denotes the type of the variable: whether it's continuous, binary from median dichotomy or non-median dichotomy.    Similar to rank correlation, the recall and precision of FOFC with tetrachoric correlation decrease as the variable goes from continuous to non-median dichotomy.  
 

\subsection{Comparing FOFC with MGM and MGM-FCI-MAX}
\subsubsection{Introduction of MGM}

MGM is an algorithm learning graphical structure over continuous and discrete variables\cite{MGM}.  Unlike FOFC, which only relies on correlation, MGM uses a pairwise graphical model\cite{mixedGraph}:
\begin{center}
    $p(x,y;\Theta)\propto exp(\sum\limits_{s=1}^{p}\sum\limits_{t=1}^{p}-\dfrac{1}{2}\beta_{st}x_s x_t+\sum\limits_{s=1}^p\alpha_sx_s+\sum\limits_{s=1}^{p}\sum\limits_{j=1}^{q}\rho_{sj}(y_j)x_s+\sum\limits_{j=1}^q\sum\limits_{r=1}^q\theta_{rj}(y_r,y_j))$
\end{center}
where there $p$ continuous variables $x$ and $q$ discrete varaibles $y$.

Using the pseudolikelihood method, MGM estimates the parameters $\Theta = [\{\beta_{st}\},\{\alpha_{s}\},\{\rho_{sj}\},\{\theta_{rj}\}]$ and use them to build an undirected graph:
\begin{itemize}
    \item an edge is added between two nodes $s$ and $t$ representing continuous variables $x_s$ and $x_t$ if $\beta_{st}\neq0$
    \item an edge is added between two nodes $s$ and $j$ representing continuous variable $x_s$ and discrete variable $y_j$ if it's not the case where $\rho_{sj}=0$ for all values of $y_j$
    \item an edge is added between two nodes $r$ and $j$ representing discrete variables $y_r$ and $y_j$ if $\theta_{rj}\neq0$ for all values of $y_j$ and $y_r$
\end{itemize}

\begin{figure}
    \centering
    \includegraphics[width=50mm]{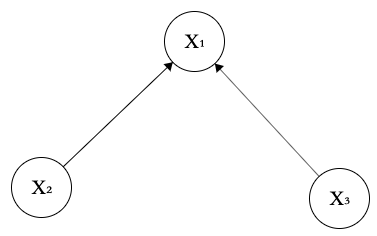}
    \caption{the True Graph with a Collider}
    \label{fig:collider}
\end{figure}

\begin{figure}
    \centering
    \includegraphics[width=50mm]{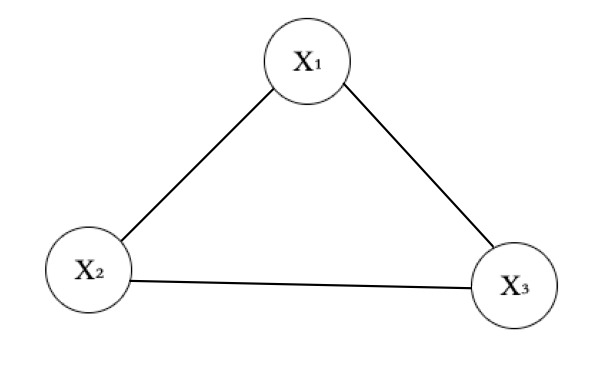}
    \caption{the Undirected Graph the MGM Will first Generate}
    \label{fig:undirect}
\end{figure}

The undirected graph is built with the idea that the absence of an edge between two nodes $x$ and $y$ means that variables $x$ and $y$ are independent conditioning on all other variables.  This undirected graph is then pruned using an independence test.  For instance, when running on a dataset with three variables where the true graph is figure \ref{fig:collider}, the MGM will first generate figure \ref{fig:undirect} since any two variables are dependent given the third one.  The final output will be a undirected graph with the edge between $X_2$ and $X_3$ removed since they are marginally independent.

\subsubsection{Customized Recall and Precision Test}

The \textit{recall} and \textit{precision} used to evaluate the FOFC are the same as before, but customized for MGM. 

MGM does not decide whether some variables share a latent common cause, but we can interpret its output in that way: if some variables form a clique, this clique is equivalent to an estimated cluster from FOFC.  Therefore, the \textit{recall} can be calculated by checking whether all variables from a true cluster form a clique in the MGM output.  Similar to the calculation for FOFC, \textit{individual recall} is calculated for each true cluster, which will be either 1 if a clique is formed or 0 if not. After this step, \textit{final recall} is calculated as the mean of \textit{individual recall}.

\begin{figure}
    \centering
    \includegraphics[width=50mm]{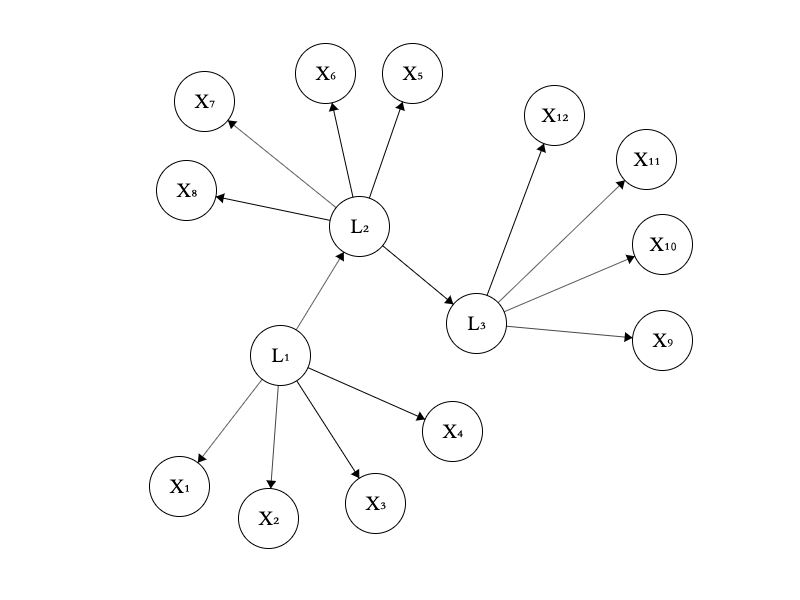}
    \caption{the Graph Used for Simulation}
    \label{fig:L3E2}
\end{figure}

The \textit{precision} for MGM is expected to measure that, for each group of variables forming a clique, how accurate it is for them to actually form a \model in the true graph.  In order to do that, the system will first check if the variables forming a \model in the true graph forms a clique in the MGM output, then will build groups of variables called \textit{fake clusters}. Each \textit{fake cluster} has the same size as each \textit{true cluster}, but variables in it do not belong to the same \model in the true graph.  For instance, given the true graph figure \ref{fig:L3E2}, the system will not only check whether variables in a  \textit{true cluster}, such as $\{X_1, X_2, X_3, X_4\}$ form a clique in the MGM output, it will also check whether variables in a \textit{fake cluster}, such as $\{X_1, X_5, X_{12}, X_2\}$, form a clique.

The \textit{precision} is calculated based on the formula of \textit{precision}:
\begin{center}
    $precision = \dfrac{tp}{tp+fp}$
\end{center}
where $tp$ stands for ``true positive" and $fp$ ``false positive". For each MGM output, the number of true clusters that are cliques in the output is treated as the $tp$ and the number of fake clusters that are cliques in the output is treated as the $fp$.  Therefore, for each MGM output, in the best case the \textit{precision} will be 1 and worse case 0.  If the MGM output is a complete graph, the \textit{precision} is 0.5 since all true and fake clusters are all cliques, i.e., $tp = fp$.

\begin{figure}
    \centering
    \includegraphics[width=\linewidth]{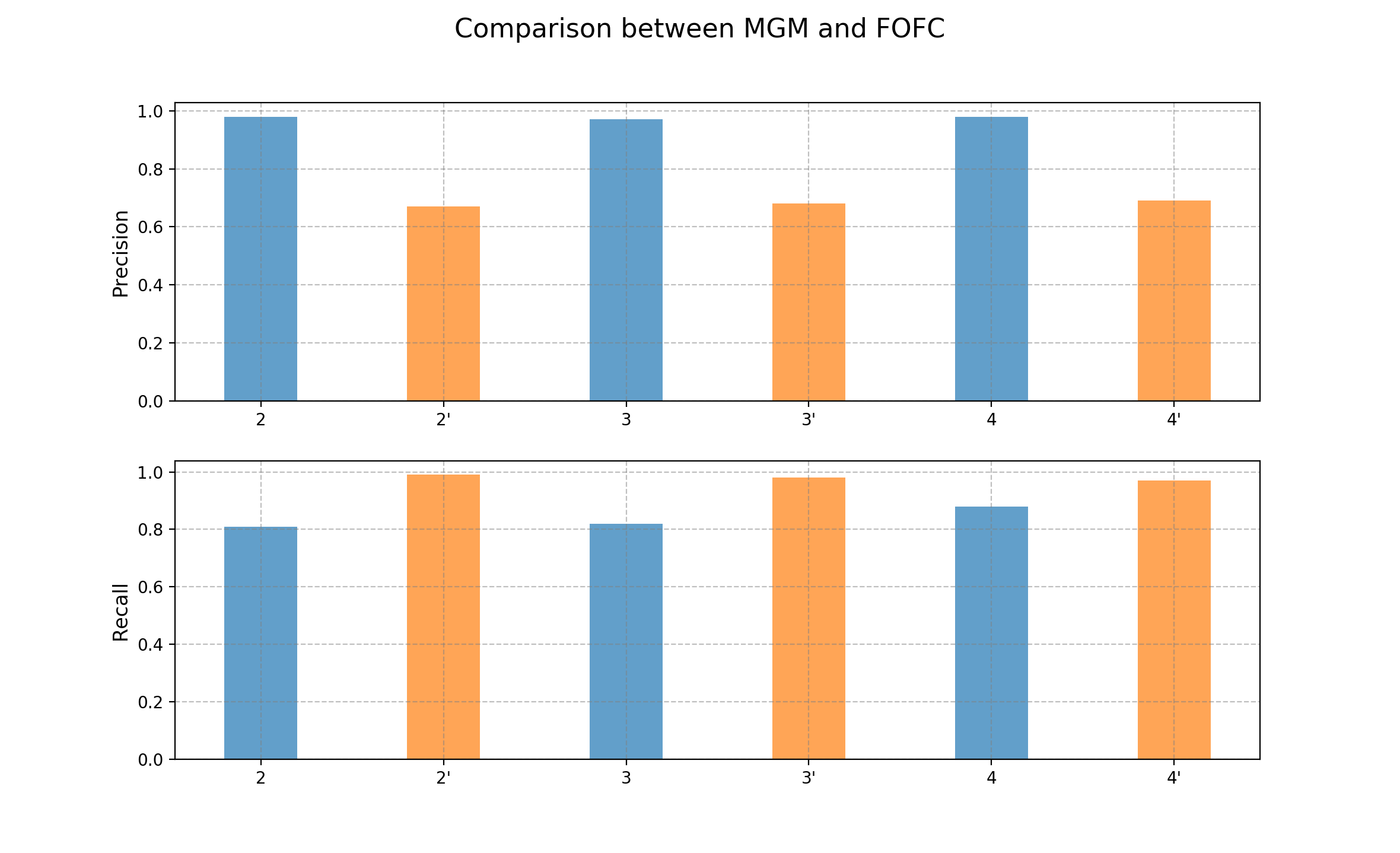}
    \caption{Comparison between FOFC (blue) and MGM (Yellow)}
    \label{fig:comp}
\end{figure}

Figure \ref{fig:L3E2} shows the graph used for simulation.  After the Gaussian data is generated, it will first be discretized with random cutoff value, then mixed with the original Gaussian data such that in each pure cluster, half of the measured variables are continuous and the other half are discrete.  Figure \ref{fig:comp} shows the result of the comparison.  For each sub-figure  the x-axis is the number of possible values the discrete variable in the mixed dataset can take: to distinguish the two algorithms, $i$ means the discrete variable can take $i$ possible values and the mixed dataset is run by FOFC;  $i'$ means $i$ possible values and is run by MGM; the y-axis is the mean of \textit{precision} (\textit{recall}) of 40 simulations with 2000 data points each time.  Note that the 40 simulations are based on the same true graph.

To make the comparison explicit, the performances of each algorithm on the same mixed dataset are put next to each other. The blue bar refers to FOFC and the orange MGM.

Figure \ref{fig:comp} shows that the \textit{recall} of FOFC and MGM are high, but MGM is better than FOFC.  It can be caused by the fact that FOFC is susceptible to the shrinking of correlation due to discretization.  The difference between the \textit{precision} of FOFC and MGM is rather obvious. 

The \textit{precision} of MGM being between 0.6 and 0.7 indicates that, when the latent variables are not independent, MGM has a hard time deciding whether variables are from the same pure cluster or not.  In fact, the \textit{precision} is consistently 0.5 when the true graph has two \modelp and the one latent variable is a direct cause of another.  If the latent variables are independent, the MGM performs as well as, or slightly better in terms of \textit{recall} than the FOFC.
\subsection{Comparing FOFC with MGM-FCI-MAX}

MGM-FCI-MAX uses the output of MGM, an undirected graph, as the input of FCI-MAX, which applies the edge orientation rule with a ``maximum probability-based search technique" \cite{Raghu2018ComparisonOS}. The setup of simulation is the same as section 4.2. Since the output of MGM-FCI-MAX is also an undirected graph (because of the setup of simulation), the \textit{precision} and \textit{recall} for MGM-FCI-MAX is calculated in the same way as for MGM.

\begin{figure}
    \centering
    \includegraphics[width=\linewidth]{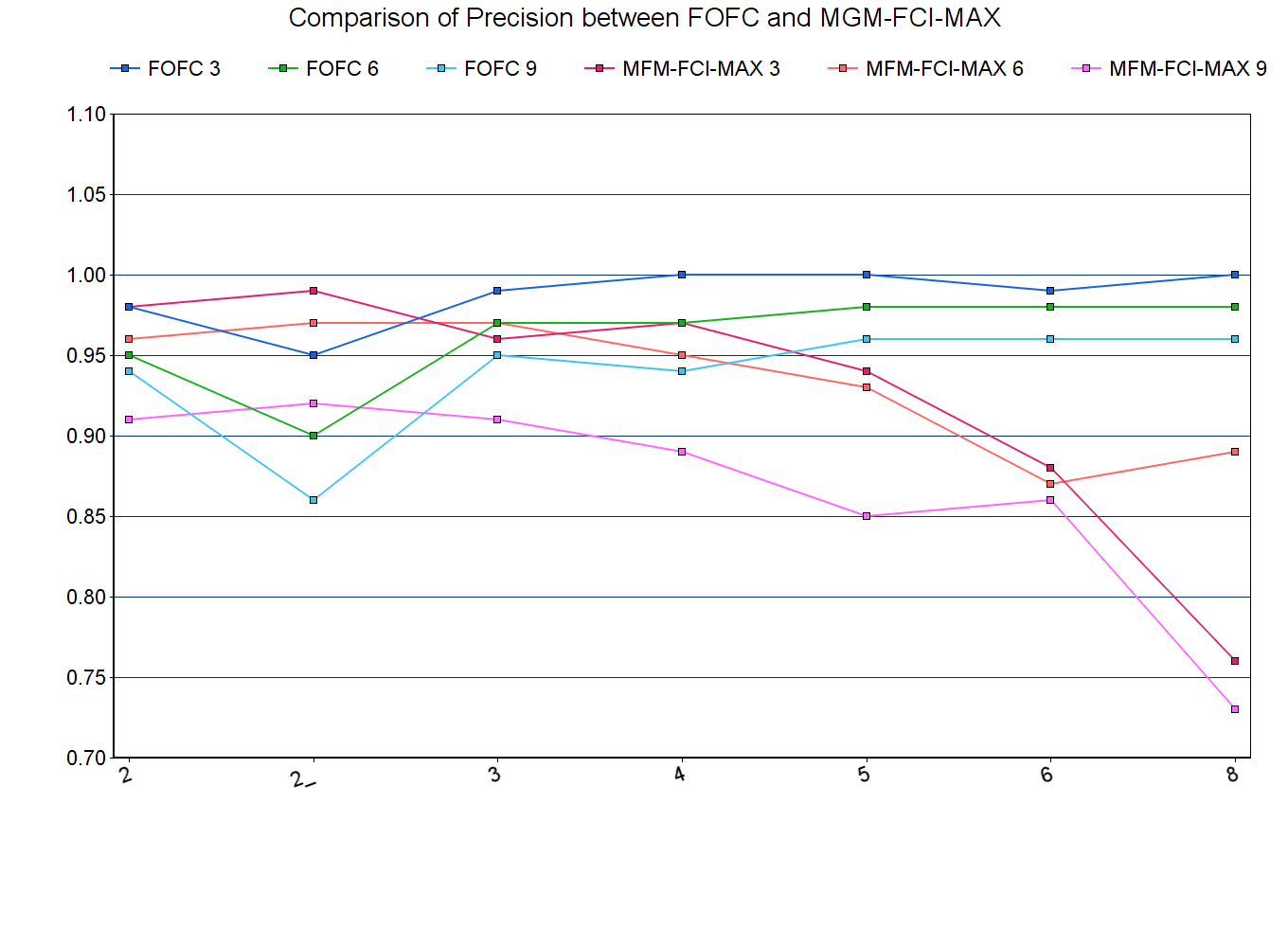}
    \caption{Precision of FOFC and MGM-FCI-MAX}
    \label{fig:PreMFM}
\end{figure}

\begin{figure}
    \centering
    \includegraphics[width=\linewidth]{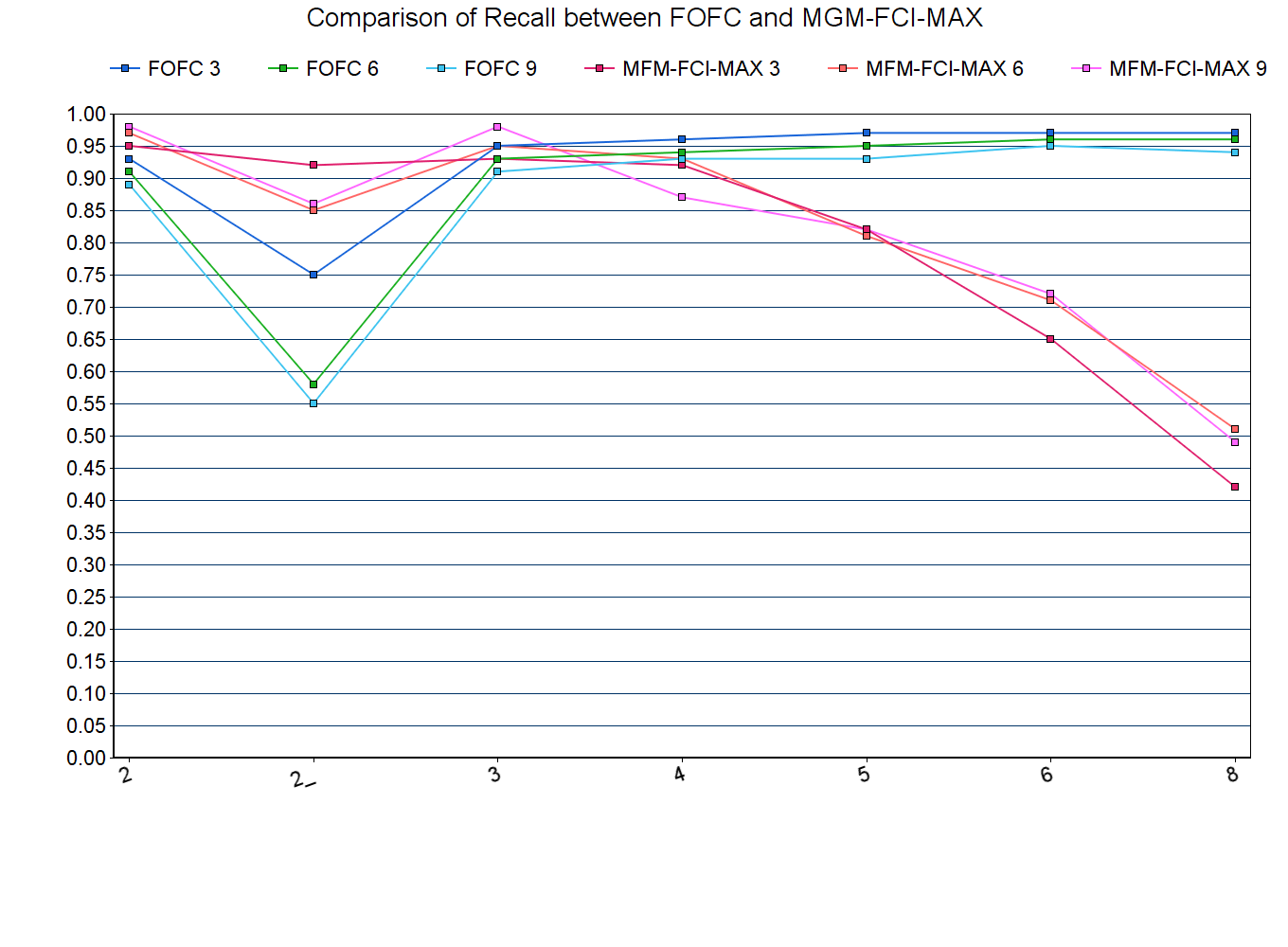}
    \caption{Recall of FOFC and MGM-FCI-MAX}
    \label{fig:RecMGM}
\end{figure}

Fig \ref{fig:PreMFM} and Fig \ref{fig:RecMGM} show that the \textit{precision} and \textit{recall} of MGM-FCI-MAX is higher than FOFC only when the discrete variables in the dataset are binary.  The \textit{precision} of MGM-FCI-MAX decreases as the number of categories the discrete variables has increases.  It can be explained by the fact that the conditional independence test used in MGM-FCI-MAX is designed specifically for mixed data, while as the number of categories increases for the discretized variable, the variable behaves similar to the continuous, as shown in \ref{fig:numCatratio}.

\newpage
\onecolumn
\section{Appendix: Proof for the Remark in Section 3.4}

\begin{remark*}
For any two discrete variables $V_i$ and $V_j$ that are derived from standard Gaussian variable $X_i$ and $X_j$ with linear discretization, such that $V_i$ has k possible values and $V_j$ has g possible values, their covariance is:
\begin{center}
    $Cov(V_i, V_j)= \sum\limits_{c_i=0}^{k-2}\sum\limits_{c_j=0}^{g-2}\Psi(S_{ic_i},S_{jc_j}, \rho_{ij})-\Psi(S_{ic_i},S_{jc_j}, 0)$
\end{center}
where $\rho_{ij}$ is the correlation between $X_i$ and $X_j$ and $S_{ic_i}$ denotes the cutoff value of $V_i$ such that $V_i=c_i$ when $S_{i(c_i-1)}<X_i\leq S_{ic_i}$.
\end{remark*}
\begin{proof}
    We are going to prove the it by induction.
    
   Basic case: when k=g=2,
\begin{align}
\begin{split}
\color{white}Cov(V_i, V_j)\color{black}Cov(V_i, V_j)
& = \Psi(-S_{i0},-S_{j0}, \rho_{ij})-\Psi(-S_{i0},-S_{j0}, 0) \ \\
 &=\int_0^{\rho_{ij}}\psi (-S_{i0},-S_{j0},r)dr\ \\
 &=\int_0^{\rho_{ij}}\dfrac{1}{2\pi\sqrt{1-r^2}}exp\{\frac{S_{i0}^2-2r S_{i0} S_{j0}+S_{j0}^2}{-2(1-r^2)}\}dr\ \\
 &=\int_0^{\rho_{ij}}\psi (S_{i0},S_{j0},r)dr\ \\
 &= \Psi(S_{i0},S_{j0}, \rho_{ij})-\Psi(S_{i0},S_{j0}, 0)
 \end{split}
 \end{align}

    \color{black}\textbf{Induction Hypothesis:} assume when $k=n'$ and $g = m'$,
    
    \color{white} for a \color{black} $Cov(V_i, V_j)= \sum\limits_{{c_i}=0}^{n'-2}\sum\limits_{{c_j}=0}^{m'-2}\Psi(S_{i{c_i}},S_{j{c_j}}, \rho_{ij})-\Psi(S_{i{c_i}},S_{j{c_j}}, 0)$ \textbf{\color{white}====}\color{black} \text{*}
    
    \color{black}Now let $k=n'+1$ (shown in Figure \ref{fig:increment}) and $g = m{'}$. 
    
    Let $\mathbb{P}_{n'+1}(V_i=c_i)$ denote the probability of $V_i=c_i$ when $V_i$ has $n'+1$ categories.  Similarly, $\mathbb{P}_{n'+1}(V_i=c_i,V_j=c_j)$ denotes the joint probability of $V_i = c_i $ and $V_j=c_j$ when $V_i$ has $n'+1$ categories, and  $\mathbb{P}_{n'}(V_i=c_i,V_j=c_j)$ the joint probability of $V_i = c_i $ and $V_j=c_j$ when $V_i$ has $n'$ categories (of course, it means that the number of categories of $V_j$ remains the same).  According to the formula:
    \begin{align*}
        \begin{split}
            \color{black}Cov(V_i, V_j)&=\mathbb{E}(V_iV_j)-\mathbb{E}(V_i)\mathbb{E}(V_j)\ \\
            &= \sum\limits_{c_i=0}^{n^{'}}\sum\limits_{c_j=0}^{m^{'}-1}c_i c_j\mathbb{P}_{n'+1}(V_i=c_i,V_j=c_j)-\sum\limits_{c_i=0}^{n^{'}}\sum\limits_{c_j=0}^{m^{'}-1}c_i c_j\mathbb{P}_{n'+1}(V_i=c_i)\mathbb{P}(V_j=c_j)\ \\
            &= \sum\limits_{c_i=0}^{n^{'}-1}\sum\limits_{c_j=0}^{m^{'}-1}c_i c_j\mathbb{P}_{n'+1}(V_i=c_i,V_j=c_j)+\sum\limits_{c_j=0}^{m^{'}-1}(n'-1+1) c_j\mathbb{P}_{n'+1}(V_i=n',V_j=c_j)\ \\ &\color{white}=\color{black}-[\sum\limits_{c_i=0}^{n^{'}-1}\sum\limits_{c_j=0}^{m^{'}-1}c_i c_j\mathbb{P}_{n'+1}(V_i=c_i)\mathbb{P}(V_j=c_j)+\sum\limits_{c_j=0}^{m^{'}-1}(n'-1+1) c_j\mathbb{P}_{n'+1}(V_i=n')\mathbb{P}(V_j=c_j)]\ \\
            &= \sum\limits_{c_i=0}^{n^{'}-2}\sum\limits_{c_j=0}^{m^{'}-1}c_i c_j\mathbb{P}_{n'+1}(V_i=c_i,V_j=c_j)\ \\
            &\color{white}=\color{black}+\sum\limits_{c_j=0}^{m^{'}-1}(n'-1)c_j[\mathbb{P}_{n'+1}(V_i=n'-1,V_j=c_j)+\mathbb{P}_{n'+1}(V_i=n',V_j=c_j)]\ \\
            &\color{white}=\color{black}+\sum\limits_{c_j=0}^{m^{'}-1}c_j\mathbb{P}_{n'+1}(V_i=n',V_j=c_j)-\{\sum\limits_{c_i=0}^{n^{'}-2}\sum\limits_{c_j=0}^{m^{'}-1}c_i c_j\mathbb{P}_{n'+1}(V_i=c_i)\mathbb{P}(V_j=c_j)\ \\
            &\color{white}=\color{black}+\sum\limits_{c_j=0}^{m^{'}-1}(n'-1) c_j\mathbb{P}(V_j=c_j)[\mathbb{P}_{n'+1}(V_i=n'-1)+\mathbb{P}_{n'+1}(V_i=n')]\ \\
            &\color{white}=\color{black}+\sum\limits_{c_j=0}^{m^{'}-1} c_j\mathbb{P}_{n'+1}(V_i=n')\mathbb{P}(V_j=c_j)\}\ \\
        \end{split}
    \end{align*}
$Cov(V_i,V_j)$ can be further reformulated by analyzing the increment of number of categories of $V_i$:
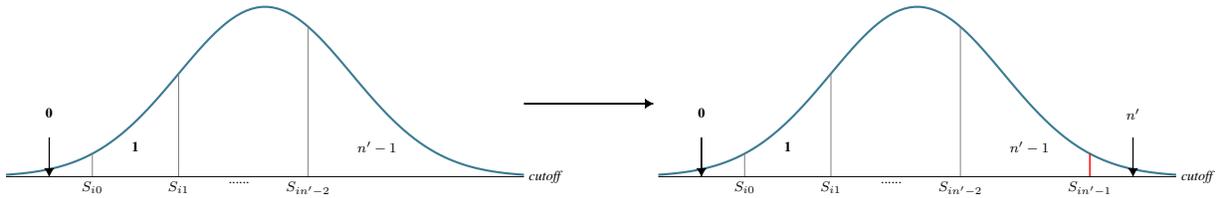
\begin{figure}[htb]
    \centering
    \resizebox{\textwidth}{!}{%
\begin{tikzpicture}
\begin{axis}[
  no markers, 
  domain=0:6, 
  samples=100,
  ymin=0,
  xlabel=\textit{cutoff},
  axis lines*=left, 
  every axis y label/.style={at=(current axis.above origin),anchor=south},
  every axis x label/.style={at=(current axis.right of origin),anchor=west},
  height=5cm, 
  width=12cm,
  xtick=\empty, 
  ytick=\empty,
  enlargelimits=false, 
  clip=false, 
  axis on top,
  grid = major,
  hide y axis
  ]

 \addplot [very thick,cyan!50!black] {gauss(x, 3, 1)};

\pgfmathsetmacro\valueA{gauss(1,3,1)}
\pgfmathsetmacro\valueB{gauss(2,3,1)}
\pgfmathsetmacro\valueC{gauss(3.5,3,1)}
\pgfmathsetmacro\valueD{gauss(5,3,1)}
\pgfmathsetmacro\valueD{gauss(4.3,3,1)}
\draw [gray]  (axis cs:1,0) -- (axis cs:1,\valueA);
\draw[gray](axis cs:2,0) -- (axis cs:2,\valueB)
    (axis cs:3.5,0) -- (axis cs:3.5,\valueC);

\draw [{Latex[ width=2mm]}-] (axis cs:0.5,0) -- (axis cs:0.5,\valueB-0.15)
    (axis cs:0.5,0) -- (axis cs:0.5,\valueB-0.15);
\draw [{Latex[ width=2mm]}-] (axis cs:7.5,\valueD) -- (axis cs:6,\valueD)
  (axis cs:7.5,\valueD) -- (axis cs:6,\valueD);

\node[below] at (axis cs:0.5, \valueB-0.07)  {$\textbf{0}$}; 
\node[below] at (axis cs:1, 0)  {$S_{i0}$}; 
\node[below] at (axis cs:1.5, \valueA+0.04)  {$\textbf{1}$}; 
\node[below] at (axis cs:2, 0)  {$S_{i1}$}; 
\node[below] at (axis cs:2.7, 0)  {......}; 
\node[below] at (axis cs:3.5, 0)  {$S_{in'-2}$};
\node[below] at (axis cs:4.3,  \valueA+0.04)  {$\textbf{$n'-1$}$}; 

\end{axis}
\end{tikzpicture}

\begin{tikzpicture}
\begin{axis}[
  no markers, 
  domain=0:6, 
  samples=100,
  ymin=0,
  axis lines*=left, 
  xlabel=\textit{cutoff},
  every axis y label/.style={at=(current axis.above origin),anchor=south},
  every axis x label/.style={at=(current axis.right of origin),anchor=west},
  height=5cm, 
  width=12cm,
  xtick=\empty, 
  ytick=\empty,
  enlargelimits=false, 
  clip=false, 
  axis on top,
  grid = major,
  hide y axis
  ]

 \addplot [very thick,cyan!50!black] {gauss(x, 3, 1)};

\pgfmathsetmacro\valueA{gauss(1,3,1)}
\pgfmathsetmacro\valueB{gauss(2,3,1)}
\pgfmathsetmacro\valueC{gauss(3.5,3,1)}
\pgfmathsetmacro\valueD{gauss(5,3,1)}

\draw [gray]  (axis cs:1,0) -- (axis cs:1,\valueA);
\draw[gray](axis cs:2,0) -- (axis cs:2,\valueB)
    (axis cs:3.5,0) -- (axis cs:3.5,\valueC);
\draw [red]  (axis cs:5,0) -- (axis cs:5,\valueD);
\draw [red]  (axis cs:5,0) -- (axis cs:5,\valueD);
\draw [{Latex[ width=2mm]}-] (axis cs:0.5,0) -- (axis cs:0.5,\valueB-0.15)
    (axis cs:0.5,0) -- (axis cs:0.5,\valueB-0.15);

\draw [{Latex[ width=2mm]}-] (axis cs:5.5,0) -- (axis cs:5.5,\valueB-0.15)
    (axis cs:5.5,0) -- (axis cs:5.5,\valueB-0.15);

\node[below] at (axis cs:0.5, \valueB-0.07)  {$\textbf{0}$}; 
\node[below] at (axis cs:1, 0)  {$S_{i0}$}; 
\node[below] at (axis cs:1.5, \valueA+0.04)  {$\textbf{1}$}; 
\node[below] at (axis cs:2, 0)  {$S_{i1}$}; 
\node[below] at (axis cs:2.7, 0)  {......}; 
\node[below] at (axis cs:3.5, 0)  {$S_{in'-2}$};
\node[below] at (axis cs:4.3,  \valueA+0.04)  {$\textbf{$n'-1$}$}; 
\node[below] at (axis cs:5, 0)  {$S_{in'-1}$};
\node[below] at (axis cs:5.5, \valueB-0.07)  {$\textbf{$n'$}$}; 

\end{axis}
\end{tikzpicture}
}
\caption{increment in number of categories for $V_i$ from $n'$ to $n'+1$}
\label{fig:increment}
\end{figure}
    \begin{enumerate}
        \item As shown in figure \ref{fig:increment}, the increment of categories does not change the former cutoff values, therefore $\mathbb{P}_{n'+1}(V_i=c_i, V_j=c_j)=\mathbb{P}_{n'}(V_i=c_i, V_j=c_j)$ when $c_i<n'-1$
        \item From figure \ref{fig:increment} it is easy to see that $\mathbb{P}_{n'+1}(V_i=n'-1, V_j=c_j)+\mathbb{P}_{n'+1}(V_i=n', V_j=c_j) = \mathbb{P}_{n'+1}(V_i\geq n'-1, V_j=c_j)=\mathbb{P}_{n'}(V_i= n'-1, V_j=c_j)$
    \end{enumerate}
\begin{align}
        \begin{split}
            Cov(V_i, V_j)&= \sum\limits_{c_i=0}^{n^{'}-2}\sum\limits_{c_j=0}^{m^{'}-1}c_i c_j\mathbb{P}_{n'+1}(V_i=c_i,V_j=c_j)\ \\
            &\color{white}=\color{black}+\sum\limits_{c_j=0}^{m^{'}-1}(n'-1)c_j[\mathbb{P}_{n'+1}(V_i=n'-1,V_j=c_j)+\mathbb{P}_{n'+1}(V_i=n',V_j=c_j)]\ \\
            &\color{white}=\color{black}+\sum\limits_{c_j=0}^{m^{'}-1}c_j\mathbb{P}_{n'+1}(V_i=c_i,V_j=c_j)\ \\ 
            &\color{white}=\color{black}-\{\sum\limits_{c_i=0}^{n^{'}-2}\sum\limits_{c_j=0}^{m^{'}-1}c_i c_j\mathbb{P}_{n'+1}(V_i=c_i)\mathbb{P}(V_j=c_j)\ \\
            &\color{white}=\color{black}+\sum\limits_{c_j=0}^{m^{'}-1}(n'-1) c_j\mathbb{P}(V_j=c_j)[\mathbb{P}_{n'+1}(V_i=n'-1)+\mathbb{P}_{n'+1}(V_i=n')]\ \\
            &\color{white}=\color{black}+\sum\limits_{c_j=0}^{m^{'}-1} c_j\mathbb{P}_{n'+1}(V_i=n')\mathbb{P}(V_j=c_j)\}\ \\
            &= \sum\limits_{c_i=0}^{n^{'}-2}\sum\limits_{c_j=0}^{m^{'}-1}c_i c_j\mathbb{P}_{n'}(V_i=c_i,V_j=c_j)+\sum\limits_{c_j=0}^{m^{'}-1}(n'-1)c_j\mathbb{P}_{n'}(V_i=n'-1,V_j=c_j)\ \\
            &\color{white}=\color{black}+\sum\limits_{c_j=0}^{m^{'}-1}c_j\mathbb{P}_{n'+1}(V_i=n',V_j=c_j)\ \\ 
            &\color{white}=\color{black}-\{\sum\limits_{c_i=0}^{n^{'}-2}\sum\limits_{c_j=0}^{m^{'}-1}c_i c_j\mathbb{P}_{n'}(V_i=c_i)\mathbb{P}(V_j=c_j)+\sum\limits_{c_j=0}^{m^{'}-1}(n'-1) c_j\mathbb{P}_{n'}(V_i=n'-1)\mathbb{P}(V_j=c_j)\ \\
            &\color{white}=\color{black}+\sum\limits_{c_j=0}^{m^{'}-1} c_j\mathbb{P}_{n'+1}(V_i=n')\mathbb{P}(V_j=c_j)\}\ \\
            &= \sum\limits_{c_i=0}^{n^{'}-1}\sum\limits_{c_j=0}^{m^{'}-1}c_i c_j\mathbb{P}_{n'}(V_i=c_i,V_j=c_j)+\sum\limits_{c_j=0}^{m^{'}-1}c_j\mathbb{P}_{n'+1}(V_i=n',V_j=c_j)\ \\
            &\color{white}=\color{black}-\sum\limits_{c_i=0}^{n^{'}-1}\sum\limits_{c_j=0}^{m^{'}-1}c_i c_j\mathbb{P}_{n'}(V_i=c_i)\mathbb{P}(V_j=c_j)-\sum\limits_{c_j=0}^{m^{'}-1} c_j\mathbb{P}_{n'+1}(V_i=n')\mathbb{P}(V_j=c_j)\ \\
            &= \sum\limits_{c_i=0}^{n^{'}-1}\sum\limits_{c_j=0}^{m^{'}-1}c_i c_j\mathbb{P}_{n'}(V_i=c_i,V_j=c_j)-\sum\limits_{c_i=0}^{n^{'}-1}\sum\limits_{c_j=0}^{m^{'}-1}c_i c_j\mathbb{P}_{n'}(V_i=c_i)\mathbb{P}(V_j=c_j)\textbf{\color{white}...............\color{black}(I)}\ \\ &\color{white}=\color{black}+\sum\limits_{c_j=0}^{m^{'}-1}c_j\mathbb{P}_{n'+1}(V_i=n',V_j=c_j)-\sum\limits_{c_j=0}^{m^{'}-1} c_j\mathbb{P}_{n'+1}(V_i=n')\mathbb{P}(V_j=c_j)\textbf{\color{white}......................\color{black}(II)}\ \\
        \end{split}
    \end{align}
    Notice that \textbf{(I)} is just $Cov(V_i,V_j)$ when $V_i$ has $n'$ categories ($k=n'$, $g=m'$); some simple calculation gives \textbf{(II)} $= \sum\limits_{c_j=0}^{m^{'}-2}\Psi(S_{in'-1},S_{jc_j}, \rho_{ij})-\Psi(S_{in'-1},S_{jc_j}, 0)$.  By Induction Hypothesis, we get:
    \begin{align*}
        \begin{split}
             &Cov(V_i, V_j)\ \\
             &=[\sum\limits_{c_i=0}^{n^{'}-2}\sum\limits_{c_j=0}^{m^{'}-2}\Psi(S_{ic_i},S_{jc_j}, \rho_{ij})-\Psi(S_{ic_i},S_{jc_j}, 0)]+[\sum\limits_{c_j=0}^{m^{'}-2}\Psi(S_{in'-1},S_{jc_j}, \rho_{ij})-\Psi(S_{in'-1},S_{jc_j}, 0)]\ \\
            &= \sum\limits_{c_i=0}^{n^{'}-1}\sum\limits_{c_j=0}^{m^{'}-2}\Psi(S_{ic_i},S_{jc_j}, \rho_{ij})-\Psi(S_{ic_i},S_{jc_j}, 0)
        \end{split}
    \end{align*}

    It can be similarly proven that $*$ holds when  $k=n'$ and $g = m'+1$.\qed
\end{proof}


%
%

\bibliographystyle{spbasic}      
\bibliography{bib}   

%
%

\end{document}